\documentclass[11pt]{article}
\usepackage{graphicx}
\usepackage{color}
\usepackage{booktabs}
\usepackage{amssymb}

\usepackage{acl}
\usepackage{times}
\usepackage{bm}
\usepackage{latexsym}
\usepackage{graphicx}
\usepackage{epstopdf}
\usepackage{algorithm}
\usepackage[noend]{algpseudocode}

\usepackage{multirow}
\usepackage{enumitem}
\usepackage{makecell}
\usepackage{amsthm}
\usepackage{amsfonts,amsmath}
\usepackage{caption}
\usepackage{subcaption}
\usepackage{url}
\usepackage[export]{adjustbox}
\usepackage{mathtools}
\usepackage{colortbl}

\usepackage[T1]{fontenc}

\usepackage[utf8]{inputenc}
\usepackage{emptypage}
\usepackage{graphicx}
\usepackage{caption}
\usepackage{adjustbox}

\usepackage{graphicx} 
\usepackage{subcaption}
\usepackage{pifont} 
\usepackage{xcolor} 
\usepackage{colortbl}
\usepackage{pgfplotstable}
\pgfplotsset{compat=1.18}
\usepackage{xcolor}

\usepackage{tabularx}
\usepackage{geometry}
\usepackage{adjustbox}
\usepackage{arydshln}
\usepackage{tcolorbox}
\usepackage{capt-of}
\usepackage{xcolor}

\usepackage{microtype}
\makeatletter
\def\BState{Sectiontate\hskip-\ALG@thistlm}
\makeatother
\title{Locate-and-Focus: Enhancing Terminology Translation in Speech Language Models}

\author{
  Suhang Wu\textsuperscript{1,4}\thanks{\,\, Work done during internship at Tongyi Lab.} \quad 
  Jialong Tang\textsuperscript{2} \quad
  Chengyi Yang\textsuperscript{1} \quad
  \textbf{Pei Zhang}\textsuperscript{2} \quad \\
  \textbf{Baosong Yang}\textsuperscript{2} \quad
  \textbf{Junhui Li}\textsuperscript{3} \quad
  \textbf{Junfeng Yao}\textsuperscript{1,4}\thanks{\,\, Corresponding authors} \quad
  \textbf{Min Zhang}\textsuperscript{3} \quad
  \textbf{Jinsong Su\textsuperscript{1,4}}\footnotemark[2]\\
  \textsuperscript{1}Department of Digital Media Technology, Xiamen University
  \textsuperscript{2}Tongyi Lab 
  \textsuperscript{3}Soochow University \\
  \textsuperscript{4}Key Laboratory of Digital Protection and Intelligent Processing of Intangible \\ Cultural Heritage of Fujian and Taiwan (Xiamen University), Ministry of Culture and Tourism, China \\
  \texttt{wusuhang@stu.xmu.edu.cn, tangjialong.tjl@alibaba-inc.com}\\
  \texttt{\{yao0010,jssu\}@xmu.edu.cn}
  \\
}

\begin{document}
\maketitle
\begin{abstract}
Direct speech translation (ST) has garnered increasing attention nowadays, yet the accurate translation of terminology within utterances remains a great challenge. In this regard, current studies mainly concentrate on leveraging various translation knowledge into ST models. However, these methods often struggle with interference from irrelevant noise and can not fully utilize the translation knowledge. To address these issues, in this paper, we propose a novel Locate-and-Focus method for terminology translation. It first effectively locates the speech clips containing terminologies within the utterance to construct translation knowledge, minimizing irrelevant information for the ST model. Subsequently, it associates the translation knowledge with the utterance and hypothesis from both audio and textual modalities, allowing the ST model to better focus on translation knowledge during translation. Experimental results across various datasets demonstrate that our method effectively locates terminologies within utterances and enhances the success rate of terminology translation, while maintaining robust general translation performance. Our code and data will be available at \url{https://github.com/DeepLearnXMU/Locate_and_Focus_ST}.
\end{abstract}

\section{Introduction}

Direct speech translation (ST) aims to convert an utterance in the source language directly into text in the target language, with recent advancements driven by the emergence of Speech Large Language Models (LLMs) \cite{DBLP:conf/interspeech/PapiTN23, DBLP:journals/corr/abs-2411-14453, peng2024surveyspeechlargelanguage,  DBLP:conf/icassp/HusseinYA0K24, DBLP:journals/csl/SethiyaM25}. 
Although significant progress has been made, dominant direct ST models still exhibit suboptimal performance in terminology translation, such as personal and drug names, which is essential for effective information delivery and professional communication \cite{DBLP:conf/taln/AilemLQ22, semenov-etal-2023-findings, DBLP:conf/wmt/BogoychevC23, DBLP:conf/emnlp/ConiaLLMP024, DBLP:conf/emnlp/YinZLM024, liu2025globalaiinclusivitylargescale}. 

\begin{figure*}[t] 
\centering 
\includegraphics[width=0.985\textwidth, height=0.290\textheight]{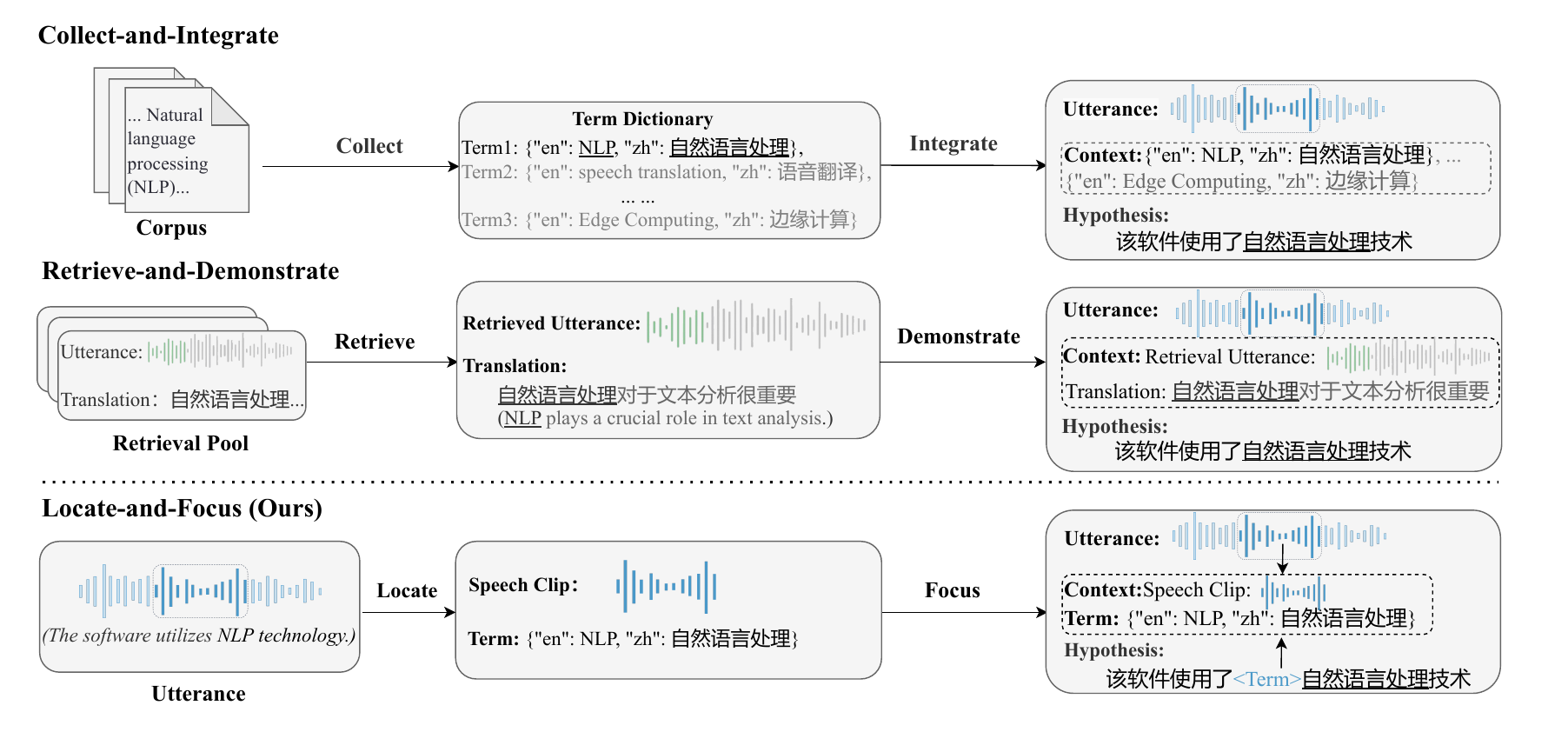} 
\caption{The differences between Locate-and-Focus and the existing paradigms. We use \textcolor{gray}{gray} to indicate information unrelated to terminology translation. Portions in the utterance and hypothesis that relate to terminology translation are highlighted in \textcolor[HTML]{4C9AC9}{blue}.}
\vspace{-0.7cm}
\label{framework_compare} 
\end{figure*}

To deal with this issue, researchers have proposed various methods that incorporate external translation knowledge. As shown in Figure \ref{framework_compare}, these methods can be roughly classified into the following two paradigms: 1) \emph{Collect-and-Integrate} \citep{DBLP:conf/icassp/GaidoTKHGI23, DBLP:conf/icassp/ChenHAHPLGBG24}. 

It collects all textual terminologies within the corpus and their translations as context to inform ST models. 2) \emph{Retrieve-and-Demonstrate} \citep{DBLP:conf/emnlp/LiLN24}. 
This paradigm employs a retriever to obtain utterance-translation pairs containing the same terms as the source utterance, 
and then provides these pairs as examples of in-context learning \citep{DBLP:conf/nips/BrownMRSKDNSSAA20}.

Despite achieving some success, the above paradigms still have two shortcomings.
On the one hand, they introduce a large amount of irrelevant information.
Specifically, the Collect-and-Integrate paradigm incorporates all corpus terminologies into the context, often including many unrelated ones such as \textit{``speech translation''} and \textit{``edge computing''}, as shown in Figure \ref{framework_compare}. The Retrieve-and-Demonstrate paradigm retrieves utterance-translation pairs that contain irrelevant sentence parts for terminology translation, such as \textit{``plays a crucial role in text analysis''}. On the other hand, due to differences in modalities or speakers, ST models struggle to fully utilize translation knowledge. Note that the Collect-and-Integrate paradigm introduces translation knowledge from the textual modality, which differs significantly from the source utterance's audio modality. Additionally, for Retrieve-and-Demonstrate, the retrieved and source utterances often originate from different speakers, with varying accents and emotions. Consequently, effectively incorporating external translation knowledge to improve terminology translation in direct ST presents significant challenges.

To tackle these challenges, we propose a novel Locate-and-Focus method for speech LLM-based terminology translation, which comprises two key steps. The terminology clip localization step employs a sliding window-based retrieval method to efficiently identify terminologies from the translation knowledge base and locate their corresponding speech clips within the utterance. This process enables the speech LLM to concentrate on portions containing terminologies, thereby reducing interference from irrelevant portions. The subsequent terminology-focused translation step associates translation knowledge with both utterances and hypotheses in both audio and textual modalities, facilitating the speech LLM to focus on translation knowledge. 
Specifically, we replace speech clips from retrieved translation knowledge with our located clips from the utterance. This process ensures that the utterance and translation knowledge share common speech clips, thereby guiding the speech LLM to focus on translation knowledge. Additionally, we encourage the speech LLM to predict a special tag before translating terminology, serving as a self-reminder to focus on the translation knowledge.

Due to the absence of terminology translation datasets for speech tasks, we collect a tailored dataset from existing ST dataset CoVoST2 \cite{DBLP:journals/corr/abs-2007-10310}, MuST-C \cite{DBLP:journals/csl/CattoniGBNT21}, and MSLT \cite{DBLP:conf/iwslt/FedermannL16, DBLP:conf/mtsummit/FedermannL17}. It contains English-to-Chinese and English-to-German translation directions. The results demonstrate that our method not only effectively locates terminologies within utterances, but also enhances the success rate of terminology translation and maintains robust general translation performance.

In summary, our contributions to this work are three-fold:
\begin{itemize}[itemsep=0pt, topsep=0pt, parsep=0pt, partopsep=0pt]
\item We propose the Locate-and-Focus method for terminology translation, which not only reduces the introduction of irrelevant information by precisely locating speech clips containing terminology, but also effectively guides speech LLMs to leverage the translation knowledge.
\item We construct a high-quality terminology translation dataset to evaluate terminology translation performance across English-to-Chinese and English-to-German translation directions.
\item Experimental results demonstrate that our method accurately locates terminologies within utterances, leading to significant improvements in terminology translation while maintaining general translation quality.
\end{itemize}

\section{Related Works}
The work related to our research encompasses the following two aspects:

\textbf{Text-based Terminology Translation.}  In this context, the main methods can be broadly categorized into three types. The first category focuses on optimizing the decoding process \cite{DBLP:conf/acl/HokampL17, DBLP:conf/naacl/PostV18, DBLP:conf/naacl/HaslerGIB18}, which improves consistency via expanded search spaces or finite-state acceptors, though it often results in poor translation quality. The second approach involves modifications to network architectures \cite{DBLP:conf/aaai/Chen0L21a, DBLP:conf/acl/WangTL22}, but significant changes in network architecture can limit its usability. Lastly, data augmentation methods include Placeholder and Code-switch. The Placeholder method replaces terminologies in both the source and target text with ordered labels, subsequently substituting these labels with the translation of terminologies after translation \cite{DBLP:journals/corr/CregoKKRYSABCDE16, DBLP:conf/coling/MichonCS20}. Code-switch method directly replaces terminologies in the source with their translation before inputting them into the model \cite{DBLP:conf/acl/DinuMFA19, DBLP:conf/eacl/BergmanisP21}. Furthermore, \citet{DBLP:conf/acl/ZhangWQSWC23} combine both Placeholder and Code-switch to achieve improved results.

Note that Placeholder and Code-switch can not be directly applied to direct ST, as replacing parts of the utterance with textual labels or translations can lead to cross-modal inconsistency. Additionally, unlike these methods that replace terminologies with labels or translations in the source text, we incorporate special tags into the model's hypothesis to improve terminology translation.

\textbf{Terminology in Speech Tasks.} 
Compared to text-based terminology translation, handling terminology in speech tasks is more complex due to the integration of more modalities \cite{DBLP:conf/icassp/HanDLCZMX22, DBLP:conf/icassp/GaidoTKHGI23, DBLP:conf/coling/LiLZ0YPQMZ024, DBLP:conf/emnlp/HuLWAZZ24, DBLP:conf/icassp/ShiYLCGZ24, DBLP:conf/icassp/ChenHAHPLGBG24}. In end-to-end automatic speech recognition (ASR), \citet{DBLP:conf/coling/LiLZ0YPQMZ024} introduce CB-Whisper, which recognizes terminology through open-vocabulary keyword spotting. \citet{DBLP:conf/emnlp/HuLWAZZ24} present VHASR, a multimodal speech recognition system. In speech translation, dominant methods can be broadly categorized into two paradigms: Collect-and-Integrate \cite{DBLP:conf/icassp/GaidoTKHGI23, DBLP:conf/icassp/ChenHAHPLGBG24} and Retrieve-and-Demonstrate \cite{DBLP:conf/emnlp/LiLN24}. As representatives of the former, \citet{DBLP:conf/icassp/GaidoTKHGI23} propose a detector to identify whether a textual terminology appears in an utterance. Similarly, \citet{DBLP:conf/icassp/ChenHAHPLGBG24} incorporate textual translations of high-frequency terminologies into prompts at a fine-grained level to aid the model in translating terminology. However, these methods do not introduce multi-modal translation knowledge. Representing the latter paradigm, \citet{DBLP:conf/emnlp/LiLN24} retrieve utterance-translation pairs and enhance terminology translation through in-context learning.

In contrast to the above studies, our work has two key advantages. First, it effectively identifies the speech clip within utterances containing terminologies, thereby reducing noise interference. Second, our method encourages the model to focus on translation knowledge from both modalities.
To the best of our knowledge, we are the first end-to-end terminology translation method that retrieves and fully utilizes multi-modal fined-granularity multi-modal fine-grained knowledge for the speech LLM.

\begin{figure*}[t] 
\centering 
\includegraphics[width=0.999\textwidth, height=0.365\textheight]{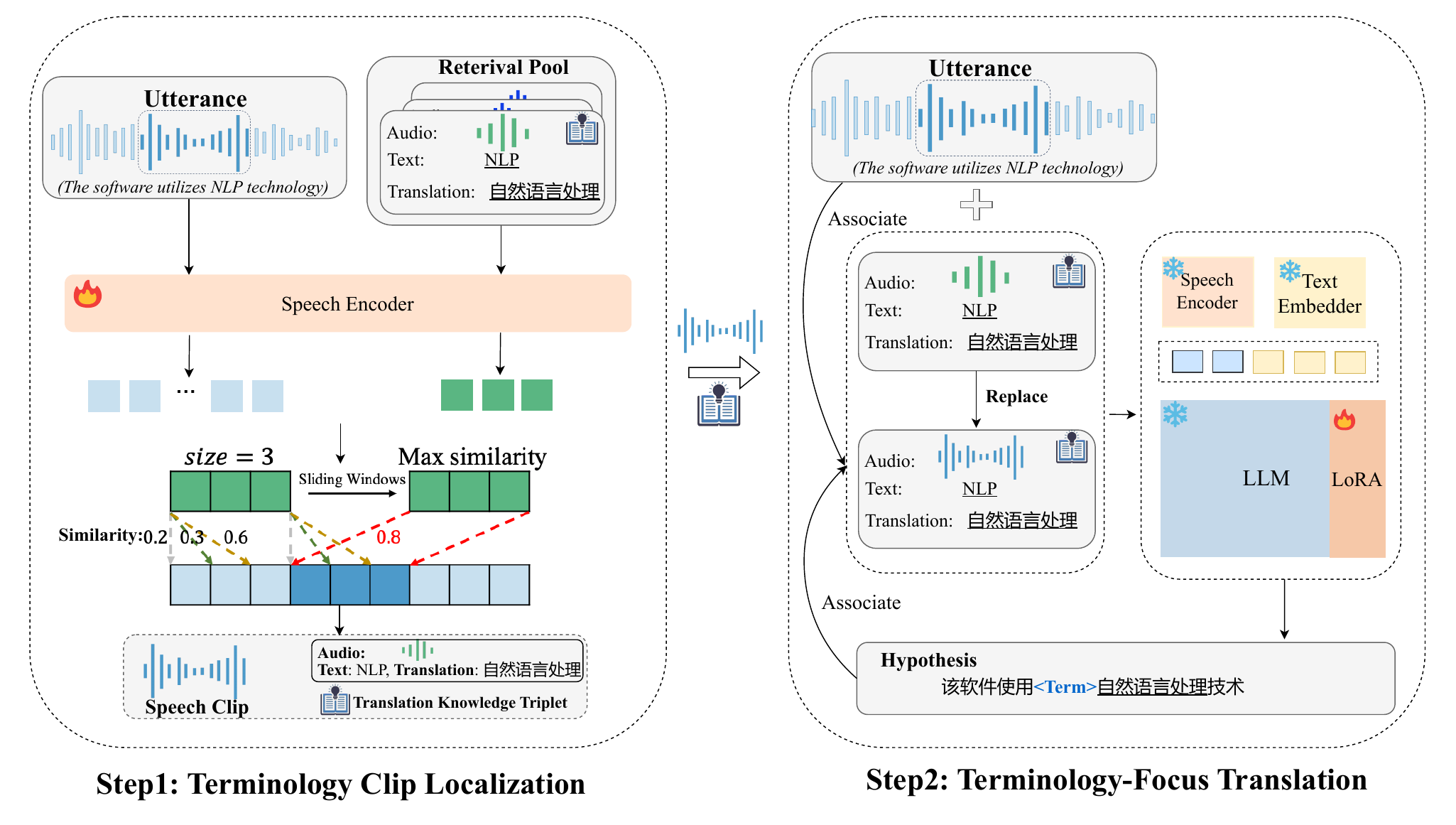} 

\caption{Overview of the \textit{Locate-and-Focus} method, which comprises the speech terminology clip localization and the terminology-focused translation steps. For a given utterance, the first step effectively identifies and locates speech clips within utterances containing the terminology. Subsequently, the second step uses audio replacement to associate the utterance and translation knowledge through their shared speech clip. It also encourages the model to predict the \texttt{<Term>} tag before translating terminology, which helps it to focus on the translation knowledge.} 
\vspace{-0.5cm}
\label{method} 
\end{figure*}

\section{Method}

In this section, we provide a detailed description of our proposed method. As shown in Figure \ref{method}, our method primarily consists of two steps: terminology clip localization and terminology-focused translation. We will elaborate on each of these steps in Sections \ref{3.1} and \ref{3.2}, followed by a discussion of the training process in Section \ref{3.3}.

\subsection{Terminology Clip Localization}
\label{3.1}

At this step, we aim to accurately retrieve terminologies within the utterance and locate their corresponding speech clips within the utterance. By locating these term-related clips in the utterance, the speech LLM can more easily focus on these key parts later, effectively minimizing irrelevant information.

Let $\mathcal{P}$ be the external translation knowledge base served as a retrieval pool, where each element is a terminology translation knowledge triplet $K=(x, c, y)$, with $x$ representing the transcript of the terminology, $c$ denoting its corresponding speech clip, and $y$ standing for its translation. Given that retrieval in the same audio modality often outperforms cross-modal retrieval \cite{DBLP:conf/emnlp/LiLN24}, we use $c$ to compute the similarity with the utterance $u$ from the test sets that require translation.

\textbf{Sliding Retrieval.} Since only certain parts of the utterance contain the terminology, it is challenging to directly calculate the similarity between $c$ and $u$ to retrieve terminologies. To address this issue, we propose a sliding window-based similarity matching method called \textit{Sliding Retrieval}, which can not only better calculate similarity but also locate the speech clip in the source utterance where the terminology is most likely to occur.

Specifically, we employ a speech encoder $SE$ to encode $c$ and $u$: $\mathbf{z}^{\text{c}} = {\rm SE}(c), \mathbf{z}^{\text{u}} = {\rm SE}(u)$, where $\mathbf{z}^{\text{c}} \!\in \mathbb{R}^{ |c|\times d}$ and $\mathbf{z}^{\text{u}} \!\in \mathbb{R}^{|u|\times d}$ represent the $d$-dimensional embeddings with lengths of $|c|$ and $|u|$, respectively. Subsequently, we utilize a sliding window with a size of \(|c|\) and a step size of $1$ to divide \(u\) into speech subsequences \([\mathbf{z}^{u}_1, \ldots, \mathbf{z}^{u}_{|u| - |c| + 1}]\)\footnote{Note that similarity calculations with different subsequences can be parallelized, resulting in only a slight increase in latency. For further details, refer to Section \ref{latency}.}. For each subsequence, we then perform max pooling on $\mathbf{z}^{u}_{i}$ and $\mathbf{z}^{\text{c}}$, followed by calculating their cosine similarity. The maximum similarity obtained will represent the similarity between $u$ and $c$, indicating the likelihood of the terminology $c$ occurring within $u$. This process is formally defined as:
\setlength{\abovedisplayskip}{2pt}
\setlength{\belowdisplayskip}{2pt}
\begin{align}
\begin{split}
\text{sim}(&u, c) = \\&\max_{i}\{ \text{Cosine}(\text{MaxPool}(\mathbf{z}^{c}), 
\text{MaxPool}(\mathbf{z}^{u}_{i})) \}
\end{split}
\end{align}

Note that we compute the similarity scores for all translation knowledge triplets in the knowledge base and then select the top-$k$ triplets with the highest scores as those whose terminology is most likely present in the utterance. Meanwhile, we identify the speech subsequence exhibiting maximum similarity and denote its corresponding speech clip as $s$, which likely contains the terminology.

\subsection{Terminology-Focused Translation}
\label{3.2}

In this step, we develop two strategies to associate the translation knowledge with the utterance and hypothesis from both audio and textual modalities, allowing the speech LLM to better focus on translation knowledge.

\paragraph{Audio Replacement.} As shown in Figure \ref{method}, we first replace the speech clip $c$ in the retrieved translation knowledge triplet $K = (x, c, y)$ with the located speech clip $s$, resulting in a new translation knowledge triplet $K^{'} = (x, s, y)$. This replacement creates an anchor that enables the utterance and translation knowledge to share identical acoustic features. When the speech LLM encounters this anchor while processing the utterance, it can more effectively focus on the relevant translation knowledge. We then provide this new triplet as the additional context along with the utterance $u$ to construct an instruction input into the speech LLM.

\paragraph{Tag Cue.} To further enhance terminology translation, we introduce special tags that serve as cues, establishing connections between the model's hypotheses and translation knowledge. Specifically, we modify the reference of training data by adding a special tag \texttt{<Term>} before the translation of each terminology. As shown in Figure \ref{method}, since ``\emph{NLP}'' is a terminology, the reference ``\emph{The software utilizes NLP technology}'' will be modified as ``\emph{The software integrates} \texttt{<Term>} \emph{NLP technology}''. Subsequently, we use these modified training data to train the speech LLM in an autoregressive manner. In this way, when the speech LLM predicts \texttt{<Term>} during inference, it cues the speech LLM to focus on the external translation knowledge triplet \( K' \) for accurate terminology translation. 

\vspace{-0.2cm}
\subsection{Training}
\label{3.3}
Note that without prior training, our terminology clip localization step can produce unsatisfactory speech clips, which may subsequently undermine the terminology-focused translation step. Therefore, we train the two steps sequentially.

The objective of training the terminology clip localization step is to ensure that $SE$ aligns with our Sliding Retrieval method. To achieve this, we employ contrastive learning for $SE$ training. Formally, our training objective \(\mathcal{L_{\text{SE}}}\) is to maximize the similarity with the positive examples while minimizing the similarity with the negative examples:
\setlength{\abovedisplayskip}{5pt}
\setlength{\belowdisplayskip}{5pt}
\begin{align}
    \mathcal{L_{\text{SE}}} &= -\log \frac{e^{\text{sim}(u, c^{+})}}{e^{\text{sim}(u, c^{+})} + \sum_{i=1}^{n} e^{\text{sim}(u, c_{i}^{-})}},
\end{align}
where \(c^{+}\) denotes the speech clip of the terminology appearing in \(u\), considered a positive example, while \(c_{i}^{-}\) denotes the \(i\)-th randomly sampled terminology speech clip, regarded as a negative example.

Subsequently, we train the model to terminology-focused translation, ensuring it effectively utilizes the provided translation knowledge during translation. Following previous studies \cite{Rajaa_SpeechLLM_Multi-Modal_LLM, DBLP:conf/icassp/ChenHAHPLGBG24}, we apply LoRA \cite{DBLP:conf/iclr/HuSWALWWC22} for fine-tuning. Formally, we train the speech LLM using the standard next token prediction loss as follows:
\setlength{\abovedisplayskip}{5pt}
\setlength{\belowdisplayskip}{5pt}
\begin{align}
\mathcal{L_{\text{LLM}}} = -\frac{1}{N} \sum_{i=1}^{N} \log P(w_i | K',u, w_{<i}),
\end{align}
where $N$ is the total number of tokens in the translations, \( w_i \) is the target token and \( P(w_i | K',u, w_{<i}) \) is the prediction probability of \( w_i \).

\begin{table}[t!]
  \centering
  \small
  \resizebox{0.5\textwidth}{!}{
  \renewcommand{\arraystretch}{1.2}
    \begin{tabularx}{0.5\textwidth}{lXXXX} 
      \toprule 
      & \multicolumn{2}{c}{\textbf{EN $\rightarrow$ ZH}} & \multicolumn{2}{c}{\textbf{EN $\rightarrow$ DE}} \\ 
      \cmidrule(lr){2-3} \cmidrule(lr){4-5}
      \textbf{Split} & \textbf{\#utt.} & \textbf{\#term.} & \textbf{\#utt.} & \textbf{\#term.} \\
      \midrule
      CoVoST2-train & 10000 & 14191 &  10000 & 14664 \\
      CoVoST2-test & 671 & 1227 & 656 & 1270 \\
      MuST-C-test & 220 & 335 & 220 & 355 \\
      MSLT-test & 213 & 294 & 164 & 280 \\
    \bottomrule
    \end{tabularx}
  }
  \caption{Statistics of our collected dataset. \#utt. indicates the number of utterances, and \#term. represents the number of terminologies.}
  \vspace{-0.6cm}
  \label{tab:dataset_analysis}
\end{table}

\section{Data Collection}
\label{S4}

Given that current speech translation datasets often lack annotated terminology translation knowledge, we create a specialized dataset for terminology translation. To be specific, we gather data from the existing ST datasets, including CoVoST2 \cite{DBLP:journals/corr/abs-2007-10310}, MuST-C \cite{DBLP:journals/csl/CattoniGBNT21}, and MSLT \cite{DBLP:conf/iwslt/FedermannL16, DBLP:conf/mtsummit/FedermannL17}. The resulting dataset features annotated terminology translation for both English-to-Chinese and English-to-German translation directions.

To achieve this, we utilize Qwen2.5-72B-Instruct \cite{DBLP:journals/corr/abs-2409-12122} to extract parallel terminology pairs from the transcripts and translations from existing ST datasets, and then manually check the extracted pairs to ensure quality. To better support ST, we use the text-to-speech (TTS) model CosyVoice2 \cite{DBLP:journals/corr/abs-2412-10117} to generate corresponding speech clips for the terms. To guarantee the quality of the generated speech, we employ the ASR model SenseVoice \cite{DBLP:journals/corr/abs-2407-04051} to transcribe the synthesized speech clips and compare these transcriptions with the source terminology. Note that we only retain clips whose transcripts have an edit distance of 3 or less from the original terminology. After this initial filtration, we also conduct a manual review to further ensure the quality of the clips. More details about our collection process are provided in Appendix \ref{append_data}.

\begin{table*}[t!]
  \centering
  \small
  \resizebox{1.0\textwidth}{!}{
  \renewcommand{\arraystretch}{1.2}
    \begin{tabularx}{1.10\textwidth}{lXXXXXXXXXXXX} 
      \toprule 
      & \multicolumn{6}{c}{\textbf{EN $\rightarrow$ ZH}} & \multicolumn{6}{c}{\textbf{EN $\rightarrow$ DE}} \\
      \cmidrule(lr){2-7} \cmidrule(lr){8-13}
      & \multicolumn{2}{c}{\textbf{CoVoST2}} & \multicolumn{2}{c}{\textbf{MuST-C}} & \multicolumn{2}{c}{\textbf{MSLT}} & \multicolumn{2}{c}{\textbf{CoVoST2}} & \multicolumn{2}{c}{\textbf{MuST-C}} & \multicolumn{2}{c}{\textbf{MSLT}}  \\ 
      \cmidrule(lr){2-3} \cmidrule(lr){4-5} \cmidrule(lr){6-7} \cmidrule(lr){8-9} \cmidrule(lr){10-11} \cmidrule(lr){12-13} 
      & \textbf{TSR} & \textbf{BLEU} & \textbf{TSR} & \textbf{BLEU} & \textbf{TSR} & \textbf{BLEU} & \textbf{TSR} & \textbf{BLEU} & \textbf{TSR} & \textbf{BLEU} & \textbf{TSR} & \textbf{BLEU} \\
      \midrule
      Base Model & 24.12	& 35.82	& 27.61	& 25.73	& 39.80	& 31.30	& 40.38	& 26.35	& 53.24	& 14.33	& 49.72	& 18.10\\
      Translation Training & 27.30 & 40.66	& 32.68	& 27.02	& 45.24	& 31.48	& 45.52	& 29.36	& 48.31	& 20.45	& 60.79	& \textbf{19.11} \\
      \midrule
      \multicolumn{13}{c}{\textit{Oracle Knowledge Setting}}\\
      \midrule
      SALM & 76.53	& 55.97	& 69.01	& 32.10	& 68.03	& 31.81	& 85.91	& 43.64	& 76.56	& 21.15& 	72.30	& 16.16 \\
      Retrieval-and-Demonstration & 60.88	& 50.22	& 58.87	& 30.18	& 70.06	& 31.34	& 57.95	& 36.09	& 57.06	& 19.46	& 53.95	& 15.18\\
      Locate-and-Focus & \textbf{90.13}	& \textbf{58.49}	& \textbf{94.09}	& \textbf{34.52}	& \textbf{91.84}	& \textbf{33.76} & \textbf{96.35}	& \textbf{45.60}	& \textbf{87.85}	& \textbf{22.06}	& \textbf{86.33}	& 17.30 \\ \hdashline
      \textit{w/o} Audio Replacement & 89.67	& 58.37	& 90.07	& 33.43	& 91.50	& 33.25	& 93.83	& 45.20	& 87.00	& 21.20	& 85.37	& 17.07\\
      \textit{w/o} Tag Cue & 89.00	& 58.25	& 88.17	& 31.09	& 90.14	& 32.05	& 90.74	& 44.97	& 85.94	& 21.36	& 83.74	& 17.24 \\
      \textit{w/o} Replacement and Cue & 88.59	& 58.32	& 86.14	& 31.44	& 89.14	& 30.05	& 91.00	& 43.29	& 81.92	& 21.88	& 76.61	& 16.67 \\
      
      \midrule
      \multicolumn{13}{c}{\textit{End-to-End Setting}}\\
      \midrule
      SALM  & 28.20	& 39.82	& 37.18	& 27.16	& 46.40	& 30.27	& 41.17	& 31.16	& 48.31	& 15.02	& 34.17	& 8.35 \\
      Retrieval-and-Demonstration & 32.93	& 41.02	& 38.31	& 26.87	& 56.80	& 30.54	& 45.37	& 32.40	& 51.97	& 16.05	& 52.88	& 15.48\\
      Locate-and-Focus & \textbf{65.53}	& \textbf{49.30}	& \textbf{75.78}	& \textbf{31.35}	& \textbf{75.51}	& 30.58	& \textbf{77.12}	& \textbf{39.66}	& \textbf{77.40}	& \textbf{21.05}	& \textbf{72.66	}& 16.98\\ \hdashline 
      \textit{w/o} Sliding Retrieval & 58.02	& 44.82	& 72.91	& 30.72	& 72.39	& 28.10	& 71.49	& 38.98	& 75.14	& 20.92	& 70.02	& 16.35 \\
      \textit{w/o} Audio Replacement & 63.49	& 49.52	& 74.62	& 31.12	& 73.91	& \textbf{32.24}	& 75.25	& 39.21	& 77.11	& 20.60	& 71.94	& \textbf{17.05}\\
      \textit{w/o} Tag Cue & 63.73	& 48.78	& 72.91	& 30.74	& 72.95	& 30.08	& 73.28	& 39.36	& 74.62	& 20.83	& 69.98	& 16.36 \\
      \textit{w/o} Replacement and Cue & 62.95	& 48.73	& 71.26	& 30.79	& 71.76	& 30.42	& 70.54	& 37.93	& 72.98	& 20.29	& 69.86	& 16.37\\
    \bottomrule
    \end{tabularx}
  }
  \caption{Performance comparison of different methods in speech terminology translation, including variants of our method. We use bold text to indicate the best performance for each metric.}
  \vspace{-0.5cm}
  \label{tab:main_res_real}
\end{table*}

The details of our collected data are presented in Table \ref{tab:dataset_analysis}. For CoVoST2 \cite{DBLP:journals/corr/abs-2007-10310}, we collect data from both the training and test splits, whereas for MuST-C \cite{DBLP:journals/csl/CattoniGBNT21} and MSLT \cite{DBLP:conf/iwslt/FedermannL16, DBLP:conf/mtsummit/FedermannL17}, we collect data only from the test splits. Note that we only retain translation samples that containing terminologies. In the subsequent process, we use only the CoVoST2 training split for model training, while MuST-C and MSLT are used as out-of-domain test sets.

\section{Experiment}

\paragraph{Base Model} 

In our experiments, we utilize the Whisper-medium \cite{DBLP:conf/icml/RadfordKXBMS23} as the speech encoder and the Qwen2-Audio-Instruct \cite{DBLP:journals/corr/abs-2407-10759} as the speech LLM. When training the speech encoder, we use 4 negative samples per example and conduct the training over 3 epochs. To ensure the translation quality of the speech LLM, we combine the original CoVoST2 training split with the terminology translation data for training. For methods requiring external translation knowledge, we use the translation knowledge base constructed in Section \ref{S4}. For further implementation details, please refer to Appendix \ref{appendix_implementation}.

\paragraph{Baselines} 
We use the representative methods as our baselines.

\begin{itemize}[itemsep=3pt, topsep=0pt, parsep=0pt, partopsep=0pt]
\item \textbf{Translation Training.} We fine-tune the speech LLM  only using the CoVoST2 training split data to enhance its translation performance. Note that it does not use external translation knowledge during inference.
\item \textbf{SALM} \cite{DBLP:conf/icassp/ChenHAHPLGBG24}. This Collect-and-Integrate method calculates term frequencies and provides the speech LLM with a fixed number of high-frequency terms and their translations as context to help model in translation. 

\item \textbf{Retrieval-and-Demonstration} \cite{DBLP:conf/emnlp/LiLN24}. This method aims to retrieve utterance-translation pairs that share terminologies with the source utterance, using them as sentence-level translation knowledge. These pairs are then employed as in-context learning examples aids to enhance terminology translation.
\end{itemize}
\vspace{-0.20cm}
\paragraph{Setups}
In our experiments, we use two different setups to supply the speech LLM with translation knowledge: 
\begin{itemize}[itemsep=3pt, topsep=0pt, parsep=0pt, partopsep=0pt]
\item \textbf{Oracle Knowledge Setting.} In this setup, the speech LLM is directly supplied with ground truth translation knowledge without any irrelevant noise, enabling the evaluation of a terminology translation method's optimal performance under ideal conditions.
\item \textbf{End-to-End Setting.} This setup requires the speech LLM to acquire external knowledge through retrieval or statistical methods, thus evaluating the terminology translation method's capability in an end-to-end manner. For SALM, we provide the translations of the top 50 most frequent terms. For the Retrieval-and-Demonstration and our method, we provide the top-5 retrieved translation knowledge.

\end{itemize}
\vspace{-0.20cm}
\paragraph{Ablation Settings}
To investigate the effects of different factors on our method, we consider the following variants for the ablation study.

\begin{itemize}[itemsep=3pt, topsep=0pt, parsep=0pt, partopsep=0pt]

  \item \textbf{\textit{w/o} Sliding Retrieval.} In this variant, the retriever uses MaxPool to calculate the similarity between utterances and speech clips, instead of employing the Sliding Retrieval approach we proposed.
  
  \item \textbf{\textit{w/o} Audio Replacement.} This variant supplies the retrieved knowledge triplet directly to the speech LLM without replacing the TTS-generated audio with the located clip from the utterance.

  \item \textbf{\textit{w/o} Tag Cue.} We exclude the use of the special tag during training in this variant, which means the model cannot use the special tag as a cue to predict when to output term translation.

  \item \textbf{\textit{w/o} Replacement and Cue.} This variant omits both the audio replacement and the tag cue during training and inference.
\end{itemize}

\paragraph{Metrics} For evaluating the retrieval performance, we use Hits@N to assess whether the correct item is included within the top-$n$ retrieved items, where $n$ is set to 1, 5, or 10. To assess the quality of terminology translation, following previous studies \citep{semenov-etal-2023-findings, DBLP:conf/emnlp/LiLN24}, we employ BLEU \cite{DBLP:conf/acl/PapineniRWZ02} and Term Success Rate (TSR) \cite{semenov-etal-2023-findings}. Term Success Rate quantifies the proportion of terminologies accurately translated within an utterance.
\vspace{-0.2cm}
\begin{table}[t!]
  \centering
  \small
  \resizebox{0.499\textwidth}{!}{
  \renewcommand{\arraystretch}{1.1}
    \begin{tabularx}{0.67\textwidth}{lXXXXXX} 
      \toprule 
      & \multicolumn{3}{c}{\textbf{CoVoST2}} & \multicolumn{3}{c}{\textbf{MuST-C}} \\ 
      \cmidrule(lr){2-4} \cmidrule(lr){5-7}
      & \textbf{Hits@1} & \textbf{Hits@5} & \textbf{Hits@10} & \textbf{Hits@1} & \textbf{Hits@5} & \textbf{Hits@10} \\
      \midrule
      \multicolumn{7}{l}{\textit{EN $\rightarrow$ ZH}}\\
      \midrule
      MaxPool & 45.07 & 56.97 &  62.18 & 55.12 & 68.17 & 74.37 \\
      MinPool & 45.80 & 55.75 & 61.53 & 53.80 & 63.66 & 70.70 \\
      AvgPool & 22.66 & 34.80 & 40.91 & 38.87 & 54.37 & 58.87 \\
      Sliding Retrieval & \textbf{61.04} & \textbf{79.22} & \textbf{85.00} & \textbf{64.23} & \textbf{82.54} & \textbf{89.58} \\
      \midrule
      \multicolumn{7}{l}{\textit{EN $\rightarrow$ DE}}\\
      \midrule
      MaxPool & 46.08 & 56.85 &  62.00 & 56.21 & 68.64 & 72.31 \\
      MinPool & 44.41 & 55.03 & 60.81 & 54.52 & 66.67 & 71.75 \\
      AvgPool & 20.66 & 34.92 & 39.67 & 35.31 & 49.15 & 55.08 \\
      Sliding Retrieval & \textbf{58.19} & \textbf{76.32} & \textbf{84.40} & \textbf{67.51} & \textbf{87.57} & \textbf{93.79}
      \\
    \bottomrule
    \end{tabularx}
  }
  \caption{Performance of the retriever on CoVoST2 and MuST-C. Please refer to Table \ref{tab:dataset_analysis} for retrieval pool sizes.}
  \vspace{-0.3cm}
  \label{tab:retriever_performance}
\end{table}

\subsection{Main Result}
As shown in Table \ref{tab:main_res_real}, we report the performance of different methods and our variants, from which we can draw the following conclusions:

\textbf{First}, providing external translation knowledge can significantly improve the success rate in terminology translation. In the Oracle Knowledge Setting, all methods incorporating external knowledge outperform both the base model and the model enhanced by translation training. This suggests that merely enhancing translation capabilities is suboptimal for effective terminology translation. We attribute this to the long-tail distribution of terms, which makes them sparse and difficult to acquire during training. Therefore, integrating external knowledge emerges as an effective approach.

\textbf{Second}, the quality of external knowledge is crucial for accurate terminology translation. In the End-to-End Setting, for instance, the performance of SALM declines as the high-frequency terms often fail to align with those in the current utterance. Retrieval-based methods face similar issues. Due to the imperfect performance of retrievers, the Retrieval-and-Demonstrate approach also experiences a performance drop, with scores falling from 60.88 to 32.93 in the CoVoST2 English-to-Chinese dataset. Therefore, we believe that further improving retrieval performance is essential for effective terminology translation.

\textbf{Third}, Locate-and-Focus surpasses existing approaches. In the End-to-End setting on the CoVoST2 English-to-Chinese dataset, it achieves a TSR of 65.53, significantly outperforming SALM's 28.20 and Retrieve-and-Demonstrate's 32.93. Additionally, it generally achieves higher BLEU scores compared to models enhanced through Translation Training. This advantage is due to the accurate translation of key terms, which is essential for overall translation quality.

\begin{figure}[t] 
\centering 
\includegraphics[width=0.50\textwidth, height=0.216\textheight]{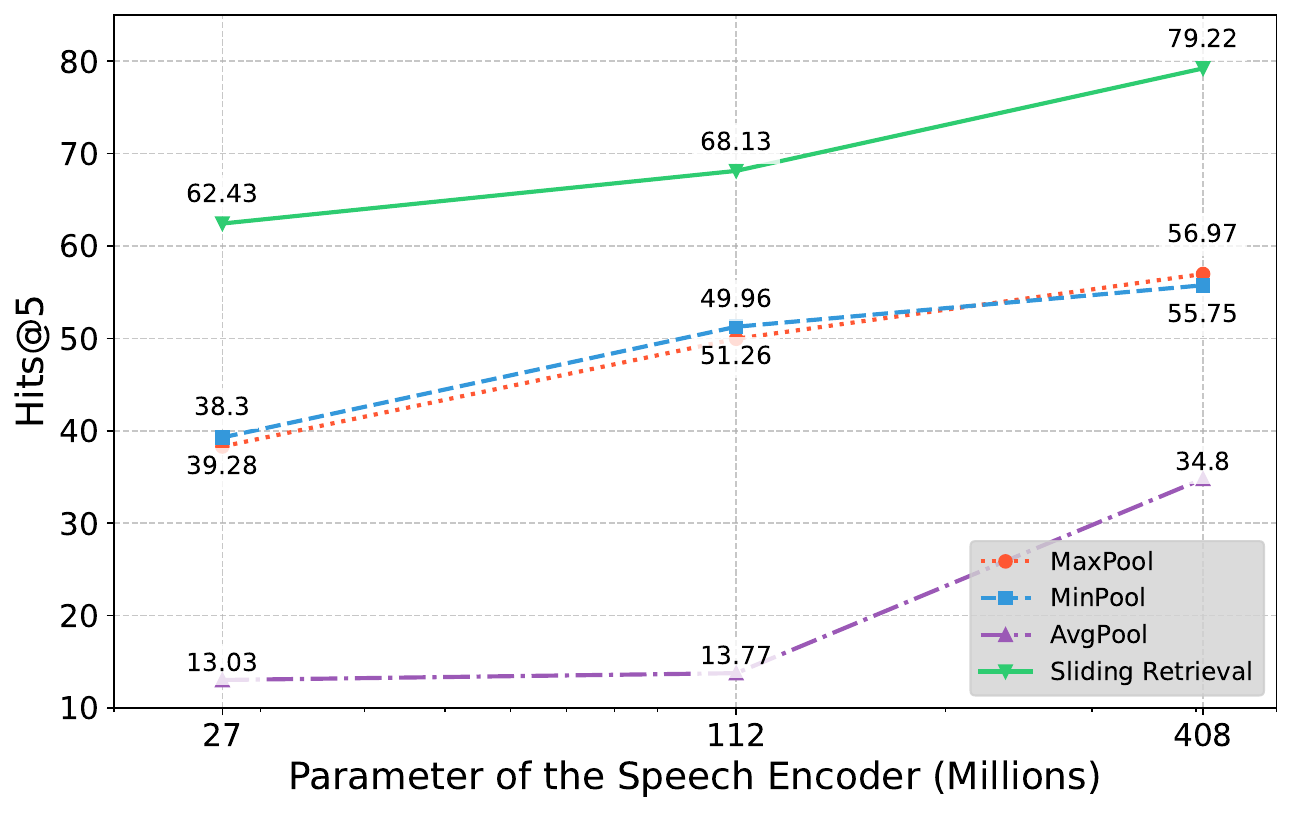} 
\caption{Comparison of Hits@5 scores for different methods using speech encoders of varying sizes.}
\vspace{-0.3cm}
\label{retriever_scaling} 
\end{figure}

\textbf{Finally}, our ablation study underscores the importance of each component in our method. We find that removing either Audio Replacement or Tag Cue results in a notable decline in performance. For example, in the Oracle knowledge setting on the MuST-C English-to-Chinese dataset, removing Audio Replacement decreases the TSR from 94.09 to 90.07, while removing the Tag Cue drops it to 88.17, and eliminating both reduces it further to 86.14. Similarly, removing Sliding Retrieval also leads to performance degradation, which we will demonstrate is related to retrieval performance.

\begin{table}[t!]
  \centering
  \small
  \resizebox{0.5\textwidth}{!}{
  \renewcommand{\arraystretch}{0.9}
    \begin{tabularx}{0.50\textwidth}{lXXXX} 
      \toprule 
      & \multicolumn{2}{c}{\textbf{CoVoST2}} & \multicolumn{2}{c}{\textbf{MuST-C}} \\ 
      \cmidrule(lr){2-3} \cmidrule(lr){4-5}
      & \textbf{TSR} & \textbf{BELU} & \textbf{TSR} & \textbf{BELU} \\
      \midrule
      \multicolumn{5}{l}{\textit{EN $\rightarrow$ ZH}}\\
      \midrule
      Top-1 &  51.67	&47.63	&58.02	&29.90 \\
      Top-5 &  \textbf{65.53}	&\textbf{49.30}	&\textbf{75.78}	&\textbf{31.35}\\
      Top-10 &  55.01	&47.57	&60.56	&29.45\\
      \midrule
      \multicolumn{5}{l}{\textit{EN $\rightarrow$ DE}}\\
      \midrule
      Top-1 &  63.89	&37.62	&69.49	&20.90\\
      Top-5 &  \textbf{77.12}	&\textbf{39.66}	&\textbf{77.40}	&\textbf{21.05}\\
      Top-10 &  69.52	&38.22	&69.77	&19.74\\
    \bottomrule
    \end{tabularx}
  }
  \caption{Performance of our method across retrieval settings, where Top-$N$ indicates the inclusion of the top $N$ highest-scoring translation knowledge triplets.}
  \vspace{-0.4cm}
  \label{tab:noise_exp}
\end{table}

\subsection{Retrieval Performance}
Given the lack of effective cross-granularity speech retrieval methods, we compare the Sliding Retrieval method with basic pooling methods as shown in Table \ref{tab:retriever_performance}, all using the same dataset to train the speech encoder. The experimental results demonstrate that our method achieves an accuracy of approximately 60\% for Hits@1 and around 85\% for Hits@10. Compared to pooling-based methods, Sliding Retrieval exhibits a significant improvement across all retrieval metrics.

To further validate our method's effectiveness across different model sizes, we conduct experiments on the English-to-Chinese subset of CoVoST2 using Whisper-base (about 27M parameters), Whisper-small (112M parameters), Whisper-medium (408M parameters) as speech encoders\footnote{We only use the encoder of Whisper and report the parameter count of the encoder.}. As shown in Figure \ref{retriever_scaling}, our method consistently achieves significantly better performance across all model sizes.

\paragraph{Quality of Located Clips} Note that Sliding Retrieval not only improves the retrieval performance, but also effectively locates the corresponding speech clips. To validate the effectiveness, we conduct a comprehensive evaluation of the located speech clips.

Using English-to-Chinese datasets, we employ Whisper-medium to locate speech clips containing ground truth terms. Human annotators are asked to subsequently verify whether these clips accurately capture the target terminology, allowing us to calculate the success rate. The results demonstrate robust performance, with terminology identification success rates of 88.10\%, 92.56\%, and 93.98\% across the CoVoST-2, MSLT, and MuST-C datasets, respectively, confirming the method's effectiveness in precise terminology localization.

\subsection{Impact of the Amount of Translation Knowledge Retrieved}
As shown in Table \ref{tab:noise_exp}, we explore the impact of providing different amounts of retrieved translation knowledge to the speech LLM on term translation performance. The results indicate that using top-1 retrieval often yields the poorest performance, while top-10 is also less effective than top-5. This is because top-1 retrieval has poor accuracy, with Hits@1 only achieving 61.04 on the English-to-Chinese CoVoST2 dataset, significantly lower than the Hits@5 score of 79.22 and Hits@10 score of 85.00, as shown in Table \ref{tab:retriever_performance}. While top-10 retrieval achieves the highest recall, it also introduces more noise from irrelevant translation knowledge. Conversely, top-5 retrieval finds a balance by providing translation knowledge with minimal noise, leading to superior performance.

\subsection{General Translation Performance}
\label{general_trans}

\begin{table}[t!]
  \centering
  \small
  \resizebox{0.5\textwidth}{!}{
  \renewcommand{\arraystretch}{1.0}
    \begin{tabular}{lcc} 
      \toprule 
       & \textbf{EN $\rightarrow$ ZH} & \textbf{EN $\rightarrow$ DE} \\ 
      \midrule
      Base Model & 38.22 & 23.61 \\
      Translation Training & \textbf{43.64} & \textbf{29.79} \\
      SALM & 43.39 & 29.47 \\
      Retrieval-and-Demonstration & 43.08 & 29.43 \\
      Locate-and-Focus & 43.48 & 29.62 \\
      \bottomrule
    \end{tabular}
  }
  \caption{Performance of methods on standard CoVoST2 test set.}
  \vspace{-0.6cm}
  \label{tab:general_trans}
\end{table}

In this section, we investigate the potential impact of enhanced terminology translation capabilities on general speech translation performance. We conduct a comprehensive evaluation on the standard CoVoST2 test sets, with BLEU scores reported in Table \ref{tab:general_trans}. The experimental results demonstrate that our approach excels not only in terminology-specific translation tasks, but also maintains robust general speech translation performance. For example, our Locate-and-Focus method achieves a BLEU score of 43.48 on the English-to-Chinese test set, approaching the performance of the Translation Training approach (43.64) while surpassing other retrieval-based methods such as SALM (43.39) and Retrieval-and-Demonstration (43.08).

\subsection{Inference Latency}
\label{latency}
Considering the critical real-time constraints of speech translation systems, we present a comprehensive evaluation of our Sliding Retrieval method's computational efficiency. Using a single NVIDIA A100 80GB GPU, we pre-compute and store speech representations generated by speech encoders, then systematically measure the time required to retrieve results from the retrieval pool using a single utterance. Our analysis encompasses 5,000 samples, with average processing times reported in Table \ref{tab:time_consume}. 

The results demonstrate that our Sliding Retrieval method introduces only negligible computational overhead compared to the MaxPool baseline. Specifically, when employing Whisper-medium, the MaxPool approach averages 0.152 ms per query, while our Sliding Retrieval method requires merely 0.217 ms, highlighting a minor difference. Note that retrieval latency is practically insignificant when considered against the 621.951 ms required by Qwen2-Audio-Instruct for the translation process. Furthermore, our analysis reveals that scaling the speech encoder parameters has minimal impact on system latency, with Sliding Retrieval averaging 0.195 ms and 0.217 ms for Whisper-base and Whisper-medium, respectively.

\begin{table}[t!]
  \centering
  \small
  \resizebox{0.49\textwidth}{!}{
  \renewcommand{\arraystretch}{1.0}
    \begin{tabularx}{0.48\textwidth}{lcc} 
        \toprule 
        \textbf{Model} & \textbf{Method} & \textbf{Time (ms)} \\
        \midrule
        \multicolumn{3}{l}{\textit{Retrieval}} \\
        \midrule
         \multirow{2}{*}{Whisper-base} & MaxPool & 0.146 \\
         & Sliding Retrieval & 0.195 \\
        \cmidrule(lr){2-3}
         \multirow{2}{*}{Whisper-small} & MaxPool & 0.150 \\
         & Sliding Retrieval & 0.214 \\
        \cmidrule(lr){2-3}
         & Sliding Retrieval & 0.217 \\
        \midrule
        \multicolumn{3}{l}{\textit{Translation}} \\
        \midrule
        Qwen2-Audio-Instruct & - & 621.951 \\
    \bottomrule
    \end{tabularx}
  }
  \caption{Time consumption of different parts in the translation process.}
  \vspace{-0.6cm}
  \label{tab:time_consume}
\end{table}
\section{Conclusion}
In this paper, we explore the critical challenge of accurately translating terminology in speech translation. We propose the \textit{Locate-and-Focus} method, which effectively minimizes noise and fully leverages translation knowledge. The method comprises two core steps: terminology clip localization and terminology-focused translation. During the first step, we identify and locate speech clips containing terminologies. Subsequently, in the terminology-focused translation step, we associate the translation knowledge with the utterance and hypothesis from both audio and textual modalities, guiding the model to focus on translation knowledge. Experimental results demonstrate that our method significantly improves terminology translation success rates across various datasets and maintains robust general translation performance. In future work, we will extend the use of terminologies to other speech tasks and investigate robust machine translation that has been widely studied in conventional NMT research \cite{DBLP:conf/emnlp/JiangLMZZHS22,DBLP:conf/coling/MiaoLKZZCWZS22}.

\section*{Limitations}
In this section, we discuss some of the main limitations of our work and how future research may be able to address them. 
\paragraph{Reliance on Predefined Terminologies} Our method depends on a predefined set of terminologies, which might not initially include all potential terms. This limitation can somewhat constrain the method's flexibility. In the future, it will be essential to explore ways to automatically construct a comprehensive and high-quality terminology knowledge base.
\paragraph{Language Coverage} Our method has been tested only on English-to-Chinese and English-to-German translations. In the future, we plan to conduct experiments in more languages to further demonstrate the method's effectiveness.
\paragraph{Exploration in Other Speech Tasks} Our method currently focuses on translation tasks, but in the future, it could be applied to other speech tasks, such as automatic speech recognition.

\section*{Acknowledgements}
The project was supported by the National Natural Science Foundation of China (No. 62036004, No. 62276219), Natural Science Foundation of Fujian Province of China (No. 2024J011001), the Public Technology Service Platform Project of Xiamen (No.3502Z20231043) and the Alibaba Research Intern Program.  We also sincerely thank the reviewers for their thoughtful and insightful comments.

\bibliography{acl25}

\begin{thebibliography}{44}
\expandafter\ifx\csname natexlab\endcsname\relax\def\natexlab#1{#1}\fi

\bibitem[{Ailem et~al.(2022)Ailem, Liu, and Qader}]{DBLP:conf/taln/AilemLQ22}
Melissa Ailem, Jingshu Liu, and Raheel Qader. 2022.
\newblock \href {https://aclanthology.org/2022.jeptalnrecital-taln.44} {Encouraging neural machine translation to satisfy terminology constraints}.
\newblock In \emph{Actes de la 29e Conf{\'{e}}rence sur le Traitement Automatique des Langues Naturelles. Volume 1 : conf{\'{e}}rence principale, {TALN-RECITAL} 2022, Avignon, France, June 27 - July 1, 2022}, page 446. {ATALA}.

\bibitem[{An et~al.(2024)An, Chen, Deng, Du, Gao, Gao, Gu, He, Hu, Hu, Ji, Li, Li, Lu, Luo, Lv, Ma, Ma, Ni, Song, Shi, Shi, Wang, Wang, Wang, Xiao, Yan, Yang, Zhang, Zhang, Zhang, Zhao, and Zheng}]{DBLP:journals/corr/abs-2407-04051}
Keyu An, Qian Chen, Chong Deng, Zhihao Du, Changfeng Gao, Zhifu Gao, Yue Gu, Ting He, Hangrui Hu, Kai Hu, Shengpeng Ji, Yabin Li, Zerui Li, Heng Lu, Haoneng Luo, Xiang Lv, Bin Ma, Ziyang Ma, Chongjia Ni, Changhe Song, Jiaqi Shi, Xian Shi, Hao Wang, Wen Wang, Yuxuan Wang, Zhangyu Xiao, Zhijie Yan, Yexin Yang, Bin Zhang, Qinglin Zhang, Shiliang Zhang, Nan Zhao, and Siqi Zheng. 2024.
\newblock \href {https://doi.org/10.48550/ARXIV.2407.04051} {Funaudiollm: Voice understanding and generation foundation models for natural interaction between humans and llms}.
\newblock \emph{CoRR}, abs/2407.04051.

\bibitem[{Bergmanis and Pinnis(2021)}]{DBLP:conf/eacl/BergmanisP21}
Toms Bergmanis and Marcis Pinnis. 2021.
\newblock \href {https://doi.org/10.18653/V1/2021.EACL-MAIN.271} {Facilitating terminology translation with target lemma annotations}.
\newblock In \emph{Proceedings of the 16th Conference of the European Chapter of the Association for Computational Linguistics: Main Volume, {EACL} 2021, Online, April 19 - 23, 2021}, pages 3105--3111. Association for Computational Linguistics.

\bibitem[{Bogoychev and Chen(2023)}]{DBLP:conf/wmt/BogoychevC23}
Nikolay Bogoychev and Pinzhen Chen. 2023.
\newblock \href {https://doi.org/10.18653/V1/2023.WMT-1.80} {Terminology-aware translation with constrained decoding and large language model prompting}.
\newblock In \emph{Proceedings of the Eighth Conference on Machine Translation, {WMT} 2023, Singapore, December 6-7, 2023}, pages 890--896. Association for Computational Linguistics.

\bibitem[{Brown et~al.(2020)Brown, Mann, Ryder, Subbiah, Kaplan, Dhariwal, Neelakantan, Shyam, Sastry, Askell, Agarwal, Herbert{-}Voss, Krueger, Henighan, Child, Ramesh, Ziegler, Wu, Winter, Hesse, Chen, Sigler, Litwin, Gray, Chess, Clark, Berner, McCandlish, Radford, Sutskever, and Amodei}]{DBLP:conf/nips/BrownMRSKDNSSAA20}
Tom~B. Brown, Benjamin Mann, Nick Ryder, Melanie Subbiah, Jared Kaplan, Prafulla Dhariwal, Arvind Neelakantan, Pranav Shyam, Girish Sastry, Amanda Askell, Sandhini Agarwal, Ariel Herbert{-}Voss, Gretchen Krueger, Tom Henighan, Rewon Child, Aditya Ramesh, Daniel~M. Ziegler, Jeffrey Wu, Clemens Winter, Christopher Hesse, Mark Chen, Eric Sigler, Mateusz Litwin, Scott Gray, Benjamin Chess, Jack Clark, Christopher Berner, Sam McCandlish, Alec Radford, Ilya Sutskever, and Dario Amodei. 2020.
\newblock \href {https://proceedings.neurips.cc/paper/2020/hash/1457c0d6bfcb4967418bfb8ac142f64a-Abstract.html} {Language models are few-shot learners}.
\newblock In \emph{Advances in Neural Information Processing Systems 33: Annual Conference on Neural Information Processing Systems 2020, NeurIPS 2020, December 6-12, 2020, virtual}.

\bibitem[{Cattoni et~al.(2021)Cattoni, Gangi, Bentivogli, Negri, and Turchi}]{DBLP:journals/csl/CattoniGBNT21}
Roldano Cattoni, Mattia Antonino~Di Gangi, Luisa Bentivogli, Matteo Negri, and Marco Turchi. 2021.
\newblock \href {https://doi.org/10.1016/J.CSL.2020.101155} {Must-c: {A} multilingual corpus for end-to-end speech translation}.
\newblock \emph{Comput. Speech Lang.}, 66:101155.

\bibitem[{Chen et~al.(2021)Chen, Chen, and Li}]{DBLP:conf/aaai/Chen0L21a}
Guanhua Chen, Yun Chen, and Victor O.~K. Li. 2021.
\newblock \href {https://doi.org/10.1609/AAAI.V35I14.17496} {Lexically constrained neural machine translation with explicit alignment guidance}.
\newblock In \emph{Thirty-Fifth {AAAI} Conference on Artificial Intelligence, {AAAI} 2021, Thirty-Third Conference on Innovative Applications of Artificial Intelligence, {IAAI} 2021, The Eleventh Symposium on Educational Advances in Artificial Intelligence, {EAAI} 2021, Virtual Event, February 2-9, 2021}, pages 12630--12638. {AAAI} Press.

\bibitem[{Chen et~al.(2024)Chen, Huang, Andrusenko, Hrinchuk, Puvvada, Li, Ghosh, Balam, and Ginsburg}]{DBLP:conf/icassp/ChenHAHPLGBG24}
Zhehuai Chen, He~Huang, Andrei Andrusenko, Oleksii Hrinchuk, Krishna~C. Puvvada, Jason Li, Subhankar Ghosh, Jagadeesh Balam, and Boris Ginsburg. 2024.
\newblock \href {https://doi.org/10.1109/ICASSP48485.2024.10447553} {{SALM:} speech-augmented language model with in-context learning for speech recognition and translation}.
\newblock In \emph{{IEEE} International Conference on Acoustics, Speech and Signal Processing, {ICASSP} 2024, Seoul, Republic of Korea, April 14-19, 2024}, pages 13521--13525. {IEEE}.

\bibitem[{Chu et~al.(2024)Chu, Xu, Yang, Wei, Wei, Guo, Leng, Lv, He, Lin, Zhou, and Zhou}]{DBLP:journals/corr/abs-2407-10759}
Yunfei Chu, Jin Xu, Qian Yang, Haojie Wei, Xipin Wei, Zhifang Guo, Yichong Leng, Yuanjun Lv, Jinzheng He, Junyang Lin, Chang Zhou, and Jingren Zhou. 2024.
\newblock \href {https://doi.org/10.48550/ARXIV.2407.10759} {Qwen2-audio technical report}.
\newblock \emph{CoRR}, abs/2407.10759.

\bibitem[{Conia et~al.(2024)Conia, Lee, Li, Minhas, Potdar, and Li}]{DBLP:conf/emnlp/ConiaLLMP024}
Simone Conia, Daniel Lee, Min Li, Umar~Farooq Minhas, Saloni Potdar, and Yunyao Li. 2024.
\newblock \href {https://aclanthology.org/2024.emnlp-main.914} {Towards cross-cultural machine translation with retrieval-augmented generation from multilingual knowledge graphs}.
\newblock In \emph{Proceedings of the 2024 Conference on Empirical Methods in Natural Language Processing, {EMNLP} 2024, Miami, FL, USA, November 12-16, 2024}, pages 16343--16360. Association for Computational Linguistics.

\bibitem[{Crego et~al.(2016)Crego, Kim, Klein, Rebollo, Yang, Senellart, Akhanov, Brunelle, Coquard, Deng, Enoue, Geiss, Johanson, Khalsa, Khiari, Ko, Kobus, Lorieux, Martins, Nguyen, Priori, Riccardi, Segal, Servan, Tiquet, Wang, Yang, Zhang, Zhou, and Zoldan}]{DBLP:journals/corr/CregoKKRYSABCDE16}
Josep~Maria Crego, Jungi Kim, Guillaume Klein, Anabel Rebollo, Kathy Yang, Jean Senellart, Egor Akhanov, Patrice Brunelle, Aur{\'{e}}lien Coquard, Yongchao Deng, Satoshi Enoue, Chiyo Geiss, Joshua Johanson, Ardas Khalsa, Raoum Khiari, Byeongil Ko, Catherine Kobus, Jean Lorieux, Leidiana Martins, Dang{-}Chuan Nguyen, Alexandra Priori, Thomas Riccardi, Natalia Segal, Christophe Servan, Cyril Tiquet, Bo~Wang, Jin Yang, Dakun Zhang, Jing Zhou, and Peter Zoldan. 2016.
\newblock \href {http://arxiv.org/abs/1610.05540} {Systran's pure neural machine translation systems}.
\newblock \emph{CoRR}, abs/1610.05540.

\bibitem[{Dinu et~al.(2019)Dinu, Mathur, Federico, and Al{-}Onaizan}]{DBLP:conf/acl/DinuMFA19}
Georgiana Dinu, Prashant Mathur, Marcello Federico, and Yaser Al{-}Onaizan. 2019.
\newblock \href {https://doi.org/10.18653/V1/P19-1294} {Training neural machine translation to apply terminology constraints}.
\newblock In \emph{Proceedings of the 57th Conference of the Association for Computational Linguistics, {ACL} 2019, Florence, Italy, July 28- August 2, 2019, Volume 1: Long Papers}, pages 3063--3068. Association for Computational Linguistics.

\bibitem[{Du et~al.(2024)Du, Wang, Chen, Shi, Lv, Zhao, Gao, Yang, Gao, Wang, Yu, Liu, Sheng, Gu, Deng, Wang, Zhang, Yan, and Zhou}]{DBLP:journals/corr/abs-2412-10117}
Zhihao Du, Yuxuan Wang, Qian Chen, Xian Shi, Xiang Lv, Tianyu Zhao, Zhifu Gao, Yexin Yang, Changfeng Gao, Hui Wang, Fan Yu, Huadai Liu, Zhengyan Sheng, Yue Gu, Chong Deng, Wen Wang, Shiliang Zhang, Zhijie Yan, and Jingren Zhou. 2024.
\newblock \href {https://doi.org/10.48550/ARXIV.2412.10117} {Cosyvoice 2: Scalable streaming speech synthesis with large language models}.
\newblock \emph{CoRR}, abs/2412.10117.

\bibitem[{Federmann and Lewis(2016)}]{DBLP:conf/iwslt/FedermannL16}
Christian Federmann and William~D. Lewis. 2016.
\newblock \href {https://aclanthology.org/2016.iwslt-1.12} {Microsoft speech language translation {(MSLT)} corpus: The {IWSLT} 2016 release for english, french and german}.
\newblock In \emph{Proceedings of the 13th International Conference on Spoken Language Translation, {IWSLT} 2016, Seattle, WA, USA, December 8-9, 2016}. International Workshop on Spoken Language Translation.

\bibitem[{Federmann and Lewis(2017)}]{DBLP:conf/mtsummit/FedermannL17}
Christian Federmann and William~D. Lewis. 2017.
\newblock \href {https://aclanthology.org/2017.mtsummit-papers.6} {The microsoft speech language translation {(MSLT)} corpus for chinese and japanese: Conversational test data for machine translation and speech recognition}.
\newblock In \emph{Proceedings of Machine Translation Summit XVI, Volume 1: Research Track, MTSummit 2017, September 18-22, 2017, Nagoya, Aichi, Japan}, pages 72--85.

\bibitem[{Gaido et~al.(2023)Gaido, Tang, Kulikov, Huang, Gong, and Inaguma}]{DBLP:conf/icassp/GaidoTKHGI23}
Marco Gaido, Yun Tang, Ilia Kulikov, Rongqing Huang, Hongyu Gong, and Hirofumi Inaguma. 2023.
\newblock \href {https://doi.org/10.1109/ICASSP49357.2023.10094689} {Named entity detection and injection for direct speech translation}.
\newblock In \emph{{IEEE} International Conference on Acoustics, Speech and Signal Processing {ICASSP} 2023, Rhodes Island, Greece, June 4-10, 2023}, pages 1--5. {IEEE}.

\bibitem[{Gupta et~al.(2024)Gupta, Dutta, and Maurya}]{DBLP:journals/corr/abs-2411-14453}
Mahendra Gupta, Maitreyee Dutta, and Chandresh~Kumar Maurya. 2024.
\newblock \href {https://doi.org/10.48550/ARXIV.2411.14453} {Direct speech-to-speech neural machine translation: {A} survey}.
\newblock \emph{CoRR}, abs/2411.14453.

\bibitem[{Han et~al.(2022)Han, Dong, Liang, Cai, Zhou, Ma, and Xu}]{DBLP:conf/icassp/HanDLCZMX22}
Minglun Han, Linhao Dong, Zhenlin Liang, Meng Cai, Shiyu Zhou, Zejun Ma, and Bo~Xu. 2022.
\newblock \href {https://doi.org/10.1109/ICASSP43922.2022.9747101} {Improving end-to-end contextual speech recognition with fine-grained contextual knowledge selection}.
\newblock In \emph{{IEEE} International Conference on Acoustics, Speech and Signal Processing, {ICASSP} 2022, Virtual and Singapore, 23-27 May 2022}, pages 8532--8536. {IEEE}.

\bibitem[{Hasler et~al.(2018)Hasler, de~Gispert, Iglesias, and Byrne}]{DBLP:conf/naacl/HaslerGIB18}
Eva Hasler, Adri{\`{a}} de~Gispert, Gonzalo Iglesias, and Bill Byrne. 2018.
\newblock \href {https://doi.org/10.18653/V1/N18-2081} {Neural machine translation decoding with terminology constraints}.
\newblock In \emph{Proceedings of the 2018 Conference of the North American Chapter of the Association for Computational Linguistics: Human Language Technologies, NAACL-HLT, New Orleans, Louisiana, USA, June 1-6, 2018, Volume 2 (Short Papers)}, pages 506--512. Association for Computational Linguistics.

\bibitem[{Hokamp and Liu(2017)}]{DBLP:conf/acl/HokampL17}
Chris Hokamp and Qun Liu. 2017.
\newblock \href {https://doi.org/10.18653/V1/P17-1141} {Lexically constrained decoding for sequence generation using grid beam search}.
\newblock In \emph{Proceedings of the 55th Annual Meeting of the Association for Computational Linguistics, {ACL} 2017, Vancouver, Canada, July 30 - August 4, Volume 1: Long Papers}, pages 1535--1546. Association for Computational Linguistics.

\bibitem[{Hu et~al.(2022)Hu, Shen, Wallis, Allen{-}Zhu, Li, Wang, Wang, and Chen}]{DBLP:conf/iclr/HuSWALWWC22}
Edward~J. Hu, Yelong Shen, Phillip Wallis, Zeyuan Allen{-}Zhu, Yuanzhi Li, Shean Wang, Lu~Wang, and Weizhu Chen. 2022.
\newblock \href {https://openreview.net/forum?id=nZeVKeeFYf9} {Lora: Low-rank adaptation of large language models}.
\newblock In \emph{The Tenth International Conference on Learning Representations, {ICLR} 2022, Virtual Event, April 25-29, 2022}. OpenReview.net.

\bibitem[{Hu et~al.(2024)Hu, Li, Wang, Ai, Zhang, and Zhao}]{DBLP:conf/emnlp/HuLWAZZ24}
Jiliang Hu, Zuchao Li, Ping Wang, Haojun Ai, Lefei Zhang, and Hai Zhao. 2024.
\newblock \href {https://aclanthology.org/2024.emnlp-main.821} {{VHASR:} {A} multimodal speech recognition system with vision hotwords}.
\newblock In \emph{Proceedings of the 2024 Conference on Empirical Methods in Natural Language Processing, {EMNLP} 2024, Miami, FL, USA, November 12-16, 2024}, pages 14791--14804. Association for Computational Linguistics.

\bibitem[{Hussein et~al.(2024)Hussein, Yan, Anastasopoulos, Watanabe, and Khudanpur}]{DBLP:conf/icassp/HusseinYA0K24}
Amir Hussein, Brian Yan, Antonios Anastasopoulos, Shinji Watanabe, and Sanjeev Khudanpur. 2024.
\newblock \href {https://doi.org/10.1109/ICASSP48485.2024.10446102} {Enhancing end-to-end conversational speech translation through target language context utilization}.
\newblock In \emph{{IEEE} International Conference on Acoustics, Speech and Signal Processing, {ICASSP} 2024, Seoul, Republic of Korea, April 14-19, 2024}, pages 11971--11975. {IEEE}.

\bibitem[{Jiang et~al.(2022)Jiang, Lu, Meng, Zhou, Zhou, Huang, and Su}]{DBLP:conf/emnlp/JiangLMZZHS22}
Hui Jiang, Ziyao Lu, Fandong Meng, Chulun Zhou, Jie Zhou, Degen Huang, and Jinsong Su. 2022.
\newblock \href {https://doi.org/10.18653/V1/2022.EMNLP-MAIN.367} {Towards robust k-nearest-neighbor machine translation}.
\newblock In \emph{Proceedings of the 2022 Conference on Empirical Methods in Natural Language Processing, {EMNLP} 2022, Abu Dhabi, United Arab Emirates, December 7-11, 2022}, pages 5468--5477. Association for Computational Linguistics.

\bibitem[{Li et~al.(2024{\natexlab{a}})Li, Liu, and Niehues}]{DBLP:conf/emnlp/LiLN24}
Siqi Li, Danni Liu, and Jan Niehues. 2024{\natexlab{a}}.
\newblock \href {https://aclanthology.org/2024.emnlp-main.708} {Optimizing rare word accuracy in direct speech translation with a retrieval-and-demonstration approach}.
\newblock In \emph{Proceedings of the 2024 Conference on Empirical Methods in Natural Language Processing, {EMNLP} 2024, Miami, FL, USA, November 12-16, 2024}, pages 12703--12719. Association for Computational Linguistics.

\bibitem[{Li et~al.(2024{\natexlab{b}})Li, Li, Zhang, Su, Yu, Piao, Qiao, Ma, Zhao, and Yang}]{DBLP:conf/coling/LiLZ0YPQMZ024}
Yuang Li, Yinglu Li, Min Zhang, Chang Su, Jiawei Yu, Mengyao Piao, Xiaosong Qiao, Miaomiao Ma, Yanqing Zhao, and Hao Yang. 2024{\natexlab{b}}.
\newblock \href {https://aclanthology.org/2024.lrec-main.262} {Cb-whisper: Contextual biasing whisper using open-vocabulary keyword-spotting}.
\newblock In \emph{Proceedings of the 2024 Joint International Conference on Computational Linguistics, Language Resources and Evaluation, {LREC/COLING} 2024, 20-25 May, 2024, Torino, Italy}, pages 2941--2946. {ELRA} and {ICCL}.

\bibitem[{Liu et~al.(2025)Liu, Ouzzani, Li, Zhang, Ou, Bouamor, Jin, and Diab}]{liu2025globalaiinclusivitylargescale}
Jiarui Liu, Iman Ouzzani, Wenkai Li, Lechen Zhang, Tianyue Ou, Houda Bouamor, Zhijing Jin, and Mona Diab. 2025.
\newblock \href {http://arxiv.org/abs/2412.18367} {Towards global ai inclusivity: A large-scale multilingual terminology dataset (gist)}.

\bibitem[{Miao et~al.(2022)Miao, Li, Kang, Zhang, Zhou, Chen, Wang, Zhang, and Su}]{DBLP:conf/coling/MiaoLKZZCWZS22}
Zhongjian Miao, Xiang Li, Liyan Kang, Wen Zhang, Chulun Zhou, Yidong Chen, Bin Wang, Min Zhang, and Jinsong Su. 2022.
\newblock \href {https://aclanthology.org/2022.coling-1.468} {Towards robust neural machine translation with iterative scheduled data-switch training}.
\newblock In \emph{Proceedings of the 29th International Conference on Computational Linguistics, {COLING} 2022, Gyeongju, Republic of Korea, October 12-17, 2022}, pages 5266--5277. International Committee on Computational Linguistics.

\bibitem[{Michon et~al.(2020)Michon, Crego, and Senellart}]{DBLP:conf/coling/MichonCS20}
Elise Michon, Josep~Maria Crego, and Jean Senellart. 2020.
\newblock \href {https://doi.org/10.18653/V1/2020.COLING-MAIN.348} {Integrating domain terminology into neural machine translation}.
\newblock In \emph{Proceedings of the 28th International Conference on Computational Linguistics, {COLING} 2020, Barcelona, Spain (Online), December 8-13, 2020}, pages 3925--3937. International Committee on Computational Linguistics.

\bibitem[{Papi et~al.(2023)Papi, Turchi, and Negri}]{DBLP:conf/interspeech/PapiTN23}
Sara Papi, Marco Turchi, and Matteo Negri. 2023.
\newblock \href {https://doi.org/10.21437/INTERSPEECH.2023-170} {Alignatt: Using attention-based audio-translation alignments as a guide for simultaneous speech translation}.
\newblock In \emph{24th Annual Conference of the International Speech Communication Association, Interspeech 2023, Dublin, Ireland, August 20-24, 2023}, pages 3974--3978. {ISCA}.

\bibitem[{Papineni et~al.(2002)Papineni, Roukos, Ward, and Zhu}]{DBLP:conf/acl/PapineniRWZ02}
Kishore Papineni, Salim Roukos, Todd Ward, and Wei{-}Jing Zhu. 2002.
\newblock \href {https://doi.org/10.3115/1073083.1073135} {Bleu: a method for automatic evaluation of machine translation}.
\newblock In \emph{Proceedings of the 40th Annual Meeting of the Association for Computational Linguistics, July 6-12, 2002, Philadelphia, PA, {USA}}, pages 311--318. {ACL}.

\bibitem[{Peng et~al.(2024)Peng, Wang, Xi, Li, Zhang, and Yu}]{peng2024surveyspeechlargelanguage}
Jing Peng, Yucheng Wang, Yu~Xi, Xu~Li, Xizhuo Zhang, and Kai Yu. 2024.
\newblock \href {http://arxiv.org/abs/2410.18908} {A survey on speech large language models}.

\bibitem[{Post and Vilar(2018)}]{DBLP:conf/naacl/PostV18}
Matt Post and David Vilar. 2018.
\newblock \href {https://doi.org/10.18653/V1/N18-1119} {Fast lexically constrained decoding with dynamic beam allocation for neural machine translation}.
\newblock In \emph{Proceedings of the 2018 Conference of the North American Chapter of the Association for Computational Linguistics: Human Language Technologies, {NAACL-HLT} 2018, New Orleans, Louisiana, USA, June 1-6, 2018, Volume 1 (Long Papers)}, pages 1314--1324. Association for Computational Linguistics.

\bibitem[{Radford et~al.(2023)Radford, Kim, Xu, Brockman, McLeavey, and Sutskever}]{DBLP:conf/icml/RadfordKXBMS23}
Alec Radford, Jong~Wook Kim, Tao Xu, Greg Brockman, Christine McLeavey, and Ilya Sutskever. 2023.
\newblock \href {https://proceedings.mlr.press/v202/radford23a.html} {Robust speech recognition via large-scale weak supervision}.
\newblock In \emph{International Conference on Machine Learning, {ICML} 2023, 23-29 July 2023, Honolulu, Hawaii, {USA}}, volume 202 of \emph{Proceedings of Machine Learning Research}, pages 28492--28518. {PMLR}.

\bibitem[{Rajaa and Tushar(2024)}]{Rajaa_SpeechLLM_Multi-Modal_LLM}
Shangeth Rajaa and Abhinav Tushar. 2024.
\newblock \href {https://github.com/skit-ai/SpeechLLM} {{SpeechLLM: Multi-Modal LLM for Speech Understanding}}.

\bibitem[{Semenov et~al.(2023)Semenov, Zouhar, Kocmi, Zhang, Zhou, and Jiang}]{semenov-etal-2023-findings}
Kirill Semenov, Vilém Zouhar, Tom Kocmi, Dongdong Zhang, Wangchunshu Zhou, and Yuchen~Eleanor Jiang. 2023.
\newblock Findings of the wmt 2023 shared task on machine translation with terminologies.
\newblock In \emph{Proceedings of the Eight Conference on Machine Translation (WMT)}. Association for Computational Linguistics.

\bibitem[{Sethiya and Maurya(2025)}]{DBLP:journals/csl/SethiyaM25}
Nivedita Sethiya and Chandresh~Kumar Maurya. 2025.
\newblock \href {https://doi.org/10.1016/J.CSL.2024.101751} {End-to-end speech-to-text translation: {A} survey}.
\newblock \emph{Comput. Speech Lang.}, 90:101751.

\bibitem[{Shi et~al.(2024)Shi, Yang, Li, Chen, Gao, and Zhang}]{DBLP:conf/icassp/ShiYLCGZ24}
Xian Shi, Yexin Yang, Zerui Li, Yanni Chen, Zhifu Gao, and Shiliang Zhang. 2024.
\newblock \href {https://doi.org/10.1109/ICASSP48485.2024.10446106} {Seaco-paraformer: {A} non-autoregressive {ASR} system with flexible and effective hotword customization ability}.
\newblock In \emph{{IEEE} International Conference on Acoustics, Speech and Signal Processing, {ICASSP} 2024, Seoul, Republic of Korea, April 14-19, 2024}, pages 10346--10350. {IEEE}.

\bibitem[{Wang et~al.(2020)Wang, Wu, and Pino}]{DBLP:journals/corr/abs-2007-10310}
Changhan Wang, Anne Wu, and Juan~Miguel Pino. 2020.
\newblock \href {http://arxiv.org/abs/2007.10310} {Covost 2: {A} massively multilingual speech-to-text translation corpus}.
\newblock \emph{CoRR}, abs/2007.10310.

\bibitem[{Wang et~al.(2022)Wang, Tan, and Liu}]{DBLP:conf/acl/WangTL22}
Shuo Wang, Zhixing Tan, and Yang Liu. 2022.
\newblock \href {https://doi.org/10.18653/V1/2022.ACL-LONG.487} {Integrating vectorized lexical constraints for neural machine translation}.
\newblock In \emph{Proceedings of the 60th Annual Meeting of the Association for Computational Linguistics (Volume 1: Long Papers), {ACL} 2022, Dublin, Ireland, May 22-27, 2022}, pages 7063--7073. Association for Computational Linguistics.

\bibitem[{Yang et~al.(2024)Yang, Zhang, Hui, Gao, Yu, Li, Liu, Tu, Zhou, Lin, Lu, Xue, Lin, Liu, Ren, and Zhang}]{DBLP:journals/corr/abs-2409-12122}
An~Yang, Beichen Zhang, Binyuan Hui, Bofei Gao, Bowen Yu, Chengpeng Li, Dayiheng Liu, Jianhong Tu, Jingren Zhou, Junyang Lin, Keming Lu, Mingfeng Xue, Runji Lin, Tianyu Liu, Xingzhang Ren, and Zhenru Zhang. 2024.
\newblock \href {https://doi.org/10.48550/ARXIV.2409.12122} {Qwen2.5-math technical report: Toward mathematical expert model via self-improvement}.
\newblock \emph{CoRR}, abs/2409.12122.

\bibitem[{Yin et~al.(2024)Yin, Zeng, Li, Meng, and Zhang}]{DBLP:conf/emnlp/YinZLM024}
Yongjing Yin, Jiali Zeng, Yafu Li, Fandong Meng, and Yue Zhang. 2024.
\newblock \href {https://aclanthology.org/2024.findings-emnlp.866} {Lexmatcher: Dictionary-centric data curation for llm-based machine translation}.
\newblock In \emph{Findings of the Association for Computational Linguistics: {EMNLP} 2024, Miami, Florida, USA, November 12-16, 2024}, pages 14767--14779. Association for Computational Linguistics.

\bibitem[{Zaratiana et~al.(2024)Zaratiana, Tomeh, Holat, and Charnois}]{DBLP:conf/naacl/ZaratianaTHC24}
Urchade Zaratiana, Nadi Tomeh, Pierre Holat, and Thierry Charnois. 2024.
\newblock \href {https://doi.org/10.18653/V1/2024.NAACL-LONG.300} {Gliner: Generalist model for named entity recognition using bidirectional transformer}.
\newblock In \emph{Proceedings of the 2024 Conference of the North American Chapter of the Association for Computational Linguistics: Human Language Technologies (Volume 1: Long Papers), {NAACL} 2024, Mexico City, Mexico, June 16-21, 2024}, pages 5364--5376. Association for Computational Linguistics.

\bibitem[{Zhang et~al.(2023)Zhang, Wang, Qin, Shi, Wang, and Chen}]{DBLP:conf/acl/ZhangWQSWC23}
Huaao Zhang, Qiang Wang, Bo~Qin, Zelin Shi, Haibo Wang, and Ming Chen. 2023.
\newblock \href {https://doi.org/10.18653/V1/2023.ACL-LONG.332} {Understanding and improving the robustness of terminology constraints in neural machine translation}.
\newblock In \emph{Proceedings of the 61st Annual Meeting of the Association for Computational Linguistics (Volume 1: Long Papers), {ACL} 2023, Toronto, Canada, July 9-14, 2023}, pages 6029--6042. Association for Computational Linguistics.

\end{thebibliography}
\bibliographystyle{acl_natbib}

\appendix
\label{sec:appendix}
\begin{table*}[!htbp]
  \centering
  \small
  \resizebox{1.00\textwidth}{!}{
  \renewcommand{\arraystretch}{1.1}
    \begin{tabularx}{1.16\textwidth}{lcccccccccc} 
      \toprule 
      & & \multicolumn{3}{c}{\textbf{CoVoST2}} & \multicolumn{3}{c}{\textbf{MuST-C}} & \multicolumn{3}{c}{\textbf{MLST}} \\ 
      \cmidrule(lr){3-5} \cmidrule(lr){6-8} \cmidrule(lr){9-11}
      \textbf{Model} & \textbf{Method} & \textbf{Hits@1} & \textbf{Hits@5} & \textbf{Hits@10} & \textbf{Hits@1} & \textbf{Hits@5} & \textbf{Hits@10}
      & \textbf{Hits@1} & \textbf{Hits@5} & \textbf{Hits@10}\\
      \midrule
      \multirow{4}{*}{Whisper-base} & MaxPool & 23.80 & 38.30 & 44.25 & 29.30 & 47.89 & 54.65 & 36.73 & 56.46 & 65.99\\
      & MinPool & 24.69 & 39.28 & 46.54 & 32.39 & 52.39 & 57.75 & 39.8 & 58.16 & 67.01 \\
      & AvgPool & 5.94 & 13.03 & 18.01 & 9.86 & 20.28 & 25.63 & 13.61 & 24.83 & 30.61 \\
      & Sliding Retrieval & \textbf{40.59} & \textbf{62.43} & \textbf{72.13} & \textbf{46.76} & \textbf{69.86} & \textbf{76.33} & \textbf{47.96} & \textbf{75.17} & \textbf{84.36} \\
      \midrule
      \multirow{4}{*}{Whisper-small} & MaxPool & 36.59 & 49.96 & 56.56 & 49.01 & 60.28 & 68.45 & 53.4 & 73.13 & 81.29 \\
      & MinPool & 37.73 & 51.26 & 58.03 & 48.45 & 62.53 & 69.85 & \textbf{56.12} & 73.81 & 81.63 \\
      & AvgPool & 6.85 & 13.77 & 17.03 & 16.62 & 27.89 & 34.65 & 14.29 & 30.27 & 37.76 \\
      & Sliding Retrieval & \textbf{45.31} & \textbf{68.13} & \textbf{78.24} & \textbf{52.96} & \textbf{76.9} & \textbf{85.92} & 55.68 & \textbf{91.15} & \textbf{94.90} \\
      \midrule
      \multirow{4}{*}{Whisper-medium} & MaxPool & 45.07 & 56.97 & 62.18 & 55.12 & 68.17 & 74.37 & 69.05 & 83.33 & 87.76 \\
      & MinPool & 45.80 & 55.75 & 61.53 & 53.80 & 63.66 & 70.70 & 61.56 & 83.00 & 87.41 \\
      & AvgPool & 22.66 & 34.80 & 40.91 & 38.87 & 54.37 & 58.87 & 46.93 & 63.27 & 70.75 \\
      & Sliding Retrieval & \textbf{61.04} & \textbf{79.22} & \textbf{85.00} & \textbf{64.23} & \textbf{82.54} & \textbf{89.58} & \textbf{71.09} & \textbf{92.86} & \textbf{97.62} \\
    \bottomrule
    \end{tabularx}
  }
  \caption{Performance of the retriever on English-to-Chinese dataset.}
  \vspace{-0.1cm}
  \label{tab:retriever_performance_appendix_enzh}
\end{table*}

\begin{table*}[!htbp]
  \centering
  \small
  \resizebox{1.00\textwidth}{!}{
  \renewcommand{\arraystretch}{1.1}
    \begin{tabularx}{1.16\textwidth}{lcccccccccc} 
      \toprule 
      & & \multicolumn{3}{c}{\textbf{CoVoST2}} & \multicolumn{3}{c}{\textbf{MuST-C}} & \multicolumn{3}{c}{\textbf{MLST}} \\ 
      \cmidrule(lr){3-5} \cmidrule(lr){6-8} \cmidrule(lr){9-11}
      \textbf{Model} & \textbf{Method} & \textbf{Hits@1} & \textbf{Hits@5} & \textbf{Hits@10} & \textbf{Hits@1} & \textbf{Hits@5} & \textbf{Hits@10}
      & \textbf{Hits@1} & \textbf{Hits@5} & \textbf{Hits@10}\\
      \midrule
      \multirow{4}{*}{Whisper-base} & MaxPool & 22.97	& 35.47	& 42.52		& 25.42	& 43.79	& 52.26	& 28.78	& 45.69	& 55.03\\
      & MinPool & 23.52	& 39.19	& 46.08		& 32.20	& 48.31	& 56.78	& 28.78	& 45.32	& 52.88 \\
      & AvgPool & 5.62	& 11.96	& 14.81		& 7.34	& 18.36	& 24.29	& 8.99	& 16.91	& 23.02 \\
      & Sliding Retrieval & \textbf{41.4}	& \textbf{61.28}	& \textbf{71.18}		& \textbf{48.02}	& \textbf{64.69}	& \textbf{76.55}	& \textbf{49.64}	& \textbf{73.74}	& \textbf{84.17} \\
      \midrule
      \multirow{4}{*}{Whisper-small} & MaxPool & 35.78	& 48.14	& 53.99		& 44.35	& 57.62	& 63.27	& 48.56	& 61.15	& 66.18 \\
      & MinPool & 37.45	& 50.12	& 55.67		& 47.17	& 59.60	& 66.10	& 48.20	& 62.23	& 69.06 \\
      & AvgPool & 5.14	& 11.95	& 14.80		& 10.45	& 18.64	& 23.44	& 11.51	& 19.78 & 26.25 \\
      & Sliding Retrieval & \textbf{44.34}	& \textbf{63.34}	& \textbf{74.82}		& \textbf{52.14}	& \textbf{79.94}	& \textbf{88.42}	& \textbf{53.67}	& \textbf{87.05}	& \textbf{92.08} \\
      \midrule
      \multirow{4}{*}{Whisper-medium} & MaxPool & 46.08	& 56.85	& 62.00		& 56.21	& 68.64	& 72.31	& 57.55	& 72.30	& 76.62 \\
      & MinPool & 44.41	& 55.03	& 60.81		& 54.52	& 66.67	& 71.75	& 53.96	& 67.99	& 74.46 \\
      & AvgPool & 20.66	& 34.92	& 39.67		& 35.31	& 49.15	& 55.08	& 37.77	& 51.08	& 58.63 \\
      & Sliding Retrieval & \textbf{58.19}	& \textbf{76.32}	& \textbf{84.40}		& \textbf{67.51}	& \textbf{87.57}	& \textbf{93.79}	& \textbf{72.66}	& \textbf{89.21}	& \textbf{93.88} \\
    \bottomrule
    \end{tabularx}
  }
  \caption{Performance of the retriever on English-to-German dataset.}
  \vspace{-0.3cm}
  \label{tab:retriever_performance_appendix_ende}
\end{table*}

\newpage
\section{Implementation Details}
\label{appendix_implementation}

\subsection{Retriever Training}
In our experiments, we utilize Whisper-medium as the primary retriever. We incorporate 4 negative samples per example and conduct training over 3 epochs, with a learning rate set at $1 \times 10^{-5}$ and a batch size of 16. This process is executable on a single NVIDIA A100 80G GPU and requires approximately 6 hours to complete.

When extracting a speech clip, we focus on the hidden state with the highest similarity. In Whisper, each hidden state represents roughly 0.02 seconds, allowing us to precisely segment the relevant portion of the speech.

\subsection{Speech LLM Training}
For fine-tuning the speech LLM, we employ the SWIFT framework \footnote{\url{https://github.com/modelscope/ms-swift}}, using LoRA with a rank of 16, an alpha of 32, and a dropout probability of 0.05. The batch size is set to 96, and the learning rate is configured at $1e-4$. We target the \texttt{q\_proj}, \texttt{k\_proj}, and \texttt{v\_proj} modules. This training procedure is executed on eight NVIDIA A100 80G GPUs and necessitates roughly 16 hours to complete.

\section{Supplementary Experimental Results}
\label{exp_all}
\begin{table*}[t!]
  \centering
  \small
  \resizebox{1.0\textwidth}{!}{
  \renewcommand{\arraystretch}{1.2}
    \begin{tabularx}{1.25\textwidth}{lXXXXXXXXXXXX} 
      \toprule 
      & \multicolumn{6}{c}{\textbf{EN $\rightarrow$ ZH}} & \multicolumn{6}{c}{\textbf{EN $\rightarrow$ DE}} \\
      \cmidrule(lr){2-7} \cmidrule(lr){8-13}
      & \multicolumn{2}{c}{\textbf{CoVoST2}} & \multicolumn{2}{c}{\textbf{MuST-C}} & \multicolumn{2}{c}{\textbf{MSLT}} & \multicolumn{2}{c}{\textbf{CoVoST2}} & \multicolumn{2}{c}{\textbf{MuST-C}} & \multicolumn{2}{c}{\textbf{MSLT}}  \\ 
      \cmidrule(lr){2-3} \cmidrule(lr){4-5} \cmidrule(lr){6-7} \cmidrule(lr){8-9} \cmidrule(lr){10-11} \cmidrule(lr){12-13} 
      & \textbf{BLEU} & \textbf{COMET} & \textbf{BLEU} & \textbf{COMET} & \textbf{BLEU} & \textbf{COMET} & \textbf{BLEU} & \textbf{COMET} & \textbf{BLEU} & \textbf{COMET} & \textbf{BLEU} & \textbf{COMET}  \\
      \midrule
      Base Model & 35.82	& 82.46	& 25.73	& 78.90	& 31.30	& \textbf{78.02}	& 26.25	& 80.61	& 14.33	& 64.97	& 18.10	& 72.93\\
      Translation Training & 40.66	& 83.23	& 27.02	& 79.26	& 31.48	& 77.99	& 29.36	& 82.26	& 20.45	& 72.42	& \textbf{19.11}	& \textbf{72.52} \\
      \midrule
      \multicolumn{13}{c}{\textit{Oracle Knowledge Setting}}\\
      \midrule
      SALM & 55.97	& 88.01	&  32.10	& 78.83	& 31.81	& 75.75	& 43.64	& 86.47	& 21.15	& 72.21	& 16.16	& 65.16 \\
      Retrieval-and-Demonstration & 50.22	& 86.57	& 30.18	& 76.65	& 31.34	& 74.93	& 36.09	& 84.56	& 19.46	& 71.42	& 15.18	& 63.93\\
      Locate-and-Focus & 58.49	& \textbf{88.78}	& \textbf{34.52}	& \textbf{79.71}	& \textbf{33.76}	& 76.62	& \textbf{45.60}	& \textbf{86.93}	& \textbf{22.06}	& \textbf{72.57}	& 17.30	& 66.70\\ 
      
      \midrule
      \multicolumn{13}{c}{\textit{End-to-End Setting}}\\
      \midrule
      SALM  & 39.82	& 76.93	& 27.16	& 74.08	& 30.27	& 72.94	& 31.16	& 81.66	& 15.02	& 67.17	& 8.35	& 5.05 \\
      Retrieval-and-Demonstration & 41.02	& 83.57	& 26.87	& 74.00	& 30.54	& 74.82	& 32.40	& 83.05	& 16.05	& 70.10	& 15.48	& 64.33\\
      Locate-and-Focus & \textbf{49.30}	& \textbf{84.51}	& \textbf{31.35}	& \textbf{77.29}	& \textbf{30.58}	& 75.90	& \textbf{39.66}	& \textbf{84.07}	& \textbf{21.05}	& \textbf{73.76}	& 16.98	& 65.32\\ 
    \bottomrule
    \end{tabularx}
  }
  \caption{Performance comparison of different methods in speech terminology translation, including variants of our method. We use bold text to indicate the best performance for each metric.}
  \label{tab:main_res_comet}
\end{table*}

\subsection{Retrieval Performance}
In Tables \ref{tab:retriever_performance_appendix_enzh} and \ref{tab:retriever_performance_appendix_ende}, we provide a detailed presentation of the performance of the Whisper-base, Whisper-small, and Whisper-medium models when employing various retrieval methods. It is evident from the results that the sliding retrieval method consistently demonstrates outstanding performance across all models and datasets examined. For example, in the CoVoST2 dataset, the sliding retrieval method applied to the Whisper-base model achieved a Hits@1 value of 40.59, surpassing the 23.80 with the MaxPool method. This significant enhancement underscores the superiority of the sliding retrieval approach. Similar trends were observed with the MuST-C and MLST datasets. These findings illustrate that sliding retrieval not only adeptly accommodates models of varying scales but also maintains robust optimization across datasets from multiple domains.

\begin{table}[t!]
  \centering
  \small
  \resizebox{0.5\textwidth}{!}{
  \renewcommand{\arraystretch}{1.0}
    \begin{tabularx}{0.60\textwidth}{lXXXXXX} 
      \toprule 
      & \multicolumn{2}{c}{\textbf{CoVoST2}} & \multicolumn{2}{c}{\textbf{MuST-C}} & \multicolumn{2}{c}{\textbf{MSLT}} \\ 
      \cmidrule(lr){2-3} \cmidrule(lr){4-5} \cmidrule(lr){6-7}
      & \textbf{TSR} & \textbf{BELU} & \textbf{TSR} & \textbf{BELU} & \textbf{TSR} & \textbf{BELU} \\
      \midrule
      \multicolumn{7}{l}{\textit{EN $\rightarrow$ ZH}}\\
      \midrule
      Top-1 &  51.67	&47.63	&58.02	&29.90 & 68.03 	&\textbf{32.29}  \\
      Top-5 &  \textbf{65.53}	&\textbf{49.30}	&\textbf{75.78}	&\textbf{31.35} & \textbf{75.51}  &	30.58 \\
      Top-10 &  55.01	&47.57	&60.56	&29.45 & 59.18 	& 24.24 \\
      \midrule
      \multicolumn{7}{l}{\textit{EN $\rightarrow$ DE}}\\
      \midrule
      Top-1 &  63.89	&37.62	&69.49	&20.90 & 68.70 	& 16.61 \\
      Top-5 &  \textbf{77.12}	&\textbf{39.66}	&\textbf{77.40}	&\textbf{21.05} & \textbf{72.66} 	& \textbf{16.98} \\
      Top-10 &  69.52	&38.22	&69.77	&19.74 & 64.38 	& 15.31 \\
    \bottomrule
    \end{tabularx}
  }
  \caption{Performance of our method in different retrieval settings. Top-$N$ represents providing the top N highest-scoring translation knowledge triplets in our retrieval setup.}
  \vspace{-0.6cm}
  \label{tab:noise_exp_appendix}
\end{table}

\subsection{Quantity of Translation Knowledge Provided}
Table \ref{tab:noise_exp_appendix} illustrates our method's performance across various translation knowledge retrieval configurations. The results demonstrate that selecting the top-5 translation knowledge entries typically yields the best performance. This highlights the importance of balancing retrieval accuracy with the minimization of irrelevant information.

For instance, in the English-to-Chinese task on the CoVoST2 dataset, providing the top-5 knowledge entries results in a TSR of 65.53 and a BELU of 49.30, outperforming both the top-1 and top-10 settings. This suggests that including more high-relevance translation options can significantly enhance accuracy and fluency. However, while the top-10 setting might seem to offer increased diversity, it often introduces unnecessary or distracting information, leading to decreased performance. This is particularly evident in the English-to-Chinese task for the MSLT dataset, where TSR and BELU drop to 59.18 and 24.24, respectively. 

\section{Additional Evaluation Metrics}
Considering that the BLEU metric is recognized to have a gap in correlation to human judgment, we supplement our evaluation with the COMET translation metric \footnote{We use wmt22-comet-da (\url{https://huggingface.co/Unbabel/wmt22-comet-da/}).}, with results shown in Table \ref{tab:main_res_comet}. The experimental results demonstrate that our method still performs well on this metric, and the trend is consistent with that observed using BLEU.

\section{Details of Data Collection}
\label{append_data}
\bigskip

\subsection{Details of Manual Annotation}
We hire three experts proficient in English and Chinese, and three proficient in English and German, to help annotate test data. Their work involves three main tasks. First, they verify whether the terms extracted by the LLM are reasonable, ensuring they are meaningful entity names and correctly translated. Each expert independently reviews the terms to prevent bias. Second, they check if the text-to-speech generated audio includes the terms, ensuring both accurate pronunciation and naturalness, and discard low-quality audio. Finally, they verify whether the audio located by our method contains the ground truth terminology, discarding any that don’t fully meet the criteria. Every sample requires agreement from three experts before retention, ensuring high quality and reliability.

\subsection{Data Sample}
\bigskip
\noindent\fbox{
    \parbox{0.97\linewidth}{
        \textbf{Instruction for Locate-and-Focus}: I've provided a selection of words along with their audio from a dictionary. You can utilize these words for the upcoming speech translations. But please note that some of them may include information unrelated to the utterance. Bilingual words: Word: ..., Audio: <audio>...</audio>, Translation: ..., ..., Word: ..., Audio: <audio>...</audio>, Translation: ... . Translate from English to Chinese: <audio>common-voice-en.mp3</audio>
    }
}

\bigskip
\noindent\fbox{
    \parbox{0.97\linewidth}{
        \textbf{Instruction for SALM}:  I've provided a selection of words from a dictionary. You can utilize these words for the upcoming speech translations. But please note that some of them may include information unrelated to the utterance. Bilingual words: Word: ..., Audio: <audio>...</audio>, Translation: .... Translate from English to Chinese: <audio>common-voice-en.mp3</audio>
    }
}
\bigskip

\noindent\fbox{
    \parbox{0.97\linewidth}{
        \textbf{Instruction for Retrieve-and-Demostration}:  I have provided a pair of sentences that include important entities. You can use these entities for the upcoming speech translations. But please note that some of them may include information unrelated to the utterance. Audio: <audio>...</audio>, Translation: ... . Translate from English to Chinese: <audio>common-voice-en.mp3</audio>
    }
}

\bigskip
\newpage
\noindent\fbox{
    \parbox{0.97\linewidth}{\textbf{Instruction for Terminology Extraction} Please meticulously extract uncommon person and entity name pairs from the provided source sentences and their corresponding translations, organizing them into a list where each pair is formatted as [term - translated term] per line. Ensure the output contains no additional text or explanations. This task requires keen attention to accurately representing terms, including names, locations, and specific domain vocabulary, to ensure that each extracted pair reflects the correct relationship between the original text and its translation.

During this process, strictly follow the output format requirements, maintaining a "A - B" structure without any extra content, to ensure clarity and precision. For clarity, consider this example: when given specific source sentences and their translations, your task is to extract and list these uncommon name pairs accurately as "Term1 - Translation1" followed by "Term2 - Translation2," and so on.

If your analysis does not uncover any name pairs that are sufficiently distinctive or significant, return "None" to indicate this outcome.
    }
}

\subsection{Types of Collected Terminology}

\begin{table}[h]
    \centering
    \begin{tabular}{lccc}
        \toprule
        Category      & CoVoST2 & Must-C & MSLT \\
        \midrule
        Person        & 313    & 191   & 129 \\
        Location      & 297    & 41    & 53  \\
        Food          & 12     & 2     & 3   \\
        Company       & 16     & 10    & 7   \\
        Biology       & 2      & 1     & 0   \\
        Organization  & 27     & 11    & 2   \\
        Health        & 3      & 1     & 0   \\
        Culture       & 22     & 1     & 2   \\
        Transport     & 13     & 4     & 0   \\
        Religion      & 62     & 7     & 0   \\
        Fashion       & 5      & 0     & 5   \\
        Science       & 2      & 3     & 2   \\
        Geography     & 9      & 0     & 2   \\
        Language      & 26     & 2     & 2   \\
        History       & 18     & 3     & 2   \\
        Politics      & 5      & 0     & 1   \\
        Architecture  & 5      & 2     & 0   \\
        Military      & 17     & 4     & 7   \\
        Environment   & 1      & 0     & 1   \\
        Education     & 29     & 4     & 3   \\
        Sport         & 2      & 0     & 5   \\
        Book          & 4      & 1     & 0   \\
        Physics       & 0      & 1     & 0   \\
        Game          & 0      & 0     & 1   \\
        Literature    & 1      & 0     & 0   \\
        Art           & 2      & 2     & 0   \\
        Music         & 2      & 0     & 1   \\
        Entertainment & 4      & 0     & 2   \\
        Award         & 5      & 3     & 1   \\
        \bottomrule
    \end{tabular}
    \caption{Terminology distribution across various categories on English-to-Chinese Data.}
    \label{tab:data_distribution_enzh}
\end{table}

\begin{table}[h]
    \centering
    \begin{tabular}{lccc}
        \toprule
        Category      & CoVoST2 & MUST-C & MSLT \\
        \midrule
        Person        & 613    & 205   & 128 \\
        Location      & 237    & 32    & 42  \\
        Food          & 10     & 1     & 8   \\
        Company       & 13     & 9     & 10  \\
        Biology       & 1      & 1     & 1   \\
        Organization  & 6      & 11    & 2   \\
        Health        & 2      & 2     & 1   \\
        Culture       & 12     & 2     & 2   \\
        Transport     & 5      & 4     & 1   \\
        Religion      & 51     & 5     & 4   \\
        Fashion       & 5      & 0     & 8   \\
        Medicine      & 0      & 2     & 0   \\
        Science       & 1      & 1     & 1   \\
        Geography     & 0      & 0     & 1   \\
        Language      & 14     & 2     & 4   \\
        History       & 11     & 3     & 1   \\
        Architecture  & 1      & 4     & 0   \\
        Military      & 11     & 1     & 4   \\
        Environment   & 0      & 0     & 1   \\
        Education     & 14     & 6     & 3   \\
        Sport         & 1      & 0     & 1   \\
        Law           & 0      & 1     & 0   \\
        Book          & 1      & 1     & 0   \\
        Game          & 1      & 0     & 0   \\
        Literature    & 1      & 0     & 0   \\
        Art           & 1      & 1     & 0   \\
        Music         & 1      & 0     & 2   \\
        Entertainment & 3      & 0     & 0   \\
        Award         & 0      & 3     & 0   \\
        \bottomrule
    \end{tabular}
    \caption{Terminology distribution across various categories on English-to-German test data.}
    \label{tab:data_distribution_ende}
\end{table}

To better analyze the terminology translation datasets we collected, we utilized the well-performing NER model GliNER-large-v2.1 \cite{DBLP:conf/naacl/ZaratianaTHC24}\footnote{\url{https://huggingface.co/urchade/gliner_large-v2.1}} to examine the types of terms present in our data. The results are presented in Table \ref{tab:data_distribution_enzh} and Table \ref{tab:data_distribution_ende}. We find some trends by comparing the data distributions in the two tables. In the English-to-Chinese and English-to-German dataset, the ``Person'' and ``Location'' categories have significantly more terms than other categories. This indicates that terms in these categories hold high importance and frequently appear in speech translation tasks. Moreover, compared to other categories, terms related to ``Food'', ``Company'', and ``Culture'' are less prevalent in both datasets, possibly because these terms are less common in typical spoken dialogues. 

\clearpage

\end{document}